\documentclass[letterpaper]{article} 
\usepackage{aaai2026}  
\usepackage{times}  
\usepackage{helvet}  
\usepackage{courier}  
\usepackage[hyphens]{url}  
\usepackage{graphicx} 
\usepackage{natbib}  
\usepackage{caption} 
\usepackage{algorithm}
\usepackage{algorithmic}
\usepackage{newfloat}
\usepackage{listings}
\DeclareCaptionStyle{ruled}{labelfont=normalfont,labelsep=colon,strut=off} 
\floatstyle{ruled}
\newfloat{listing}{tb}{lst}{}
\floatname{listing}{Listing}
\usepackage{amsmath}       
\usepackage{mathtools}
\usepackage{amsthm}
\usepackage{tcolorbox}
\usepackage{pifont}
\usepackage{makecell}
\usepackage{array}
\usepackage{subcaption}
\usepackage{diagbox}
\usepackage{multirow}  
\usepackage[accsupp]{axessibility}
\usepackage{bbding}
\usepackage{colortbl}
 \usepackage{booktabs}
\usepackage{dblfloatfix}
\nocopyright 

\usepackage{cleveref} 
\title{MinD: Learning A Dual-System World Model for Real-Time Planning \\ and Implicit Risk Analysis}

\usepackage{bibentry}

\author{%
\small
  Xiaowei Chi\textsuperscript{\rm 1,2$^{* \dagger}$}, Kuangzhi Ge\textsuperscript{\rm 3$^{*}$}, Jiaming Liu\textsuperscript{\rm 3$^{\dagger}$}, Siyuan Zhou\textsuperscript{\rm 1,2}, Peidong Jia\textsuperscript{\rm 3}, Zichen He\textsuperscript{\rm 3}, \\ Yuzhen Liu\textsuperscript{\rm 1}, Tingguang Li\textsuperscript{\rm 1}, Lei Han\textsuperscript{\rm 1},  Sirui Han\textsuperscript{\rm 2}~\textsuperscript{\Envelope}, Shanghang Zhang\textsuperscript{\rm 3}~\textsuperscript{\Envelope}, Yike Guo \vspace{0.3cm} \\
}

\affiliations{
    \small
  \textsuperscript{\rm 1} Tencent Robotics X; 
  \textsuperscript{\rm 2}  Hong Kong University of Science and Technology; 
  \textsuperscript{\rm 3} Peking University \vspace{0.3cm}\\
  $^{*}$ Equal contribution, $^{\dagger}$ Project lead, \Envelope Corresponding author\\
}
    
\begin{document}

\maketitle
\vspace{-2em}
\definecolor{DeepYellow}{rgb}{0.87, 0.72, 0.05} 
\definecolor{DeepGreen}{rgb}{0.0, 0.5, 0.0} 
\vspace{-1em}

\begin{abstract}

Recently, Video Generation Models (VGMs) have been widely used as feature extraction backbones for VLA models, leveraging internet-scale pretraining for robust dynamics modeling.
However, world models are beyond visual representation encoders; existing methods fail to tap into the potential of their powerful distribution modeling capabilities for predicting future states. 
This oversight stems from two challenges: integrating generative processes into feature learning is technically challenging and conceptually underdeveloped.
Moreover, applying generative world models in robotics is challenging—naively generating future frames through video diffusion is computationally inefficient and unsuitable for real-time control.
%
%
To address these challenges, we introduce \textbf{Manipulate in Dream (MinD)}, a dual-system world model designed for risk-aware planning that enables real-time prediction. MinD runs two asynchronous diffusion processes: a low-frequency visual generator that first creates a future scene, and a high-frequency diffusion policy that then outputs actions. 
Our key insight is that robotic policies rely not on completely denoised image frames, but instead on low-resolution latents efficiently produced in a single denoising timestep, still serving as an effective future state representation.
To connect this early prediction to actions, we propose a video-action diffusion matching module (DiffMatcher), with a novel co-training strategy that uses separate schedulers for each diffusion model. Specifically, we introduce a diffusion-forcing mechanism to DiffMatcher that aligns their intermediate representations during training, helping the fast action model better understand video-based predictions.
Our method achieves a \textbf{63\%} success rate on RL-Bench, \textbf{60\%} on real-world Franka with \textbf{11.3} FPS, demonstrating that a single-step feature is highly effective for learning control signals. As a significant secondary finding, we demonstrate that MinD can identify 74\% of potential task failures in advance by analyzing generated video clips, while the nascent feature could also serve as an indicator. This provides a valuable, real-time signal for safety monitoring and human intervention. More demo videos: \textcolor{blue}{manipulate-in-dream.github.io}

\end{abstract}

\section{Introduction}
\label{sec:intro}


Recent advances in generative modeling, particularly Video Generation Models (VGMs)~\cite{sora, agarwal2025cosmos}, have provided a powerful foundation for building world models that integrate perception and prediction from raw visual inputs. In embodied AI, these models are increasingly used as feature extraction backbones for Vision-Language-Action (VLA) policies~\cite{yang2023unisim, RoboDreamer, hu2024vpp, bu2025univla}, leveraging internet-scale pretraining for robust dynamics modeling.

\begin{figure}[t]
    \centering
    \includegraphics[width=0.95\linewidth ]{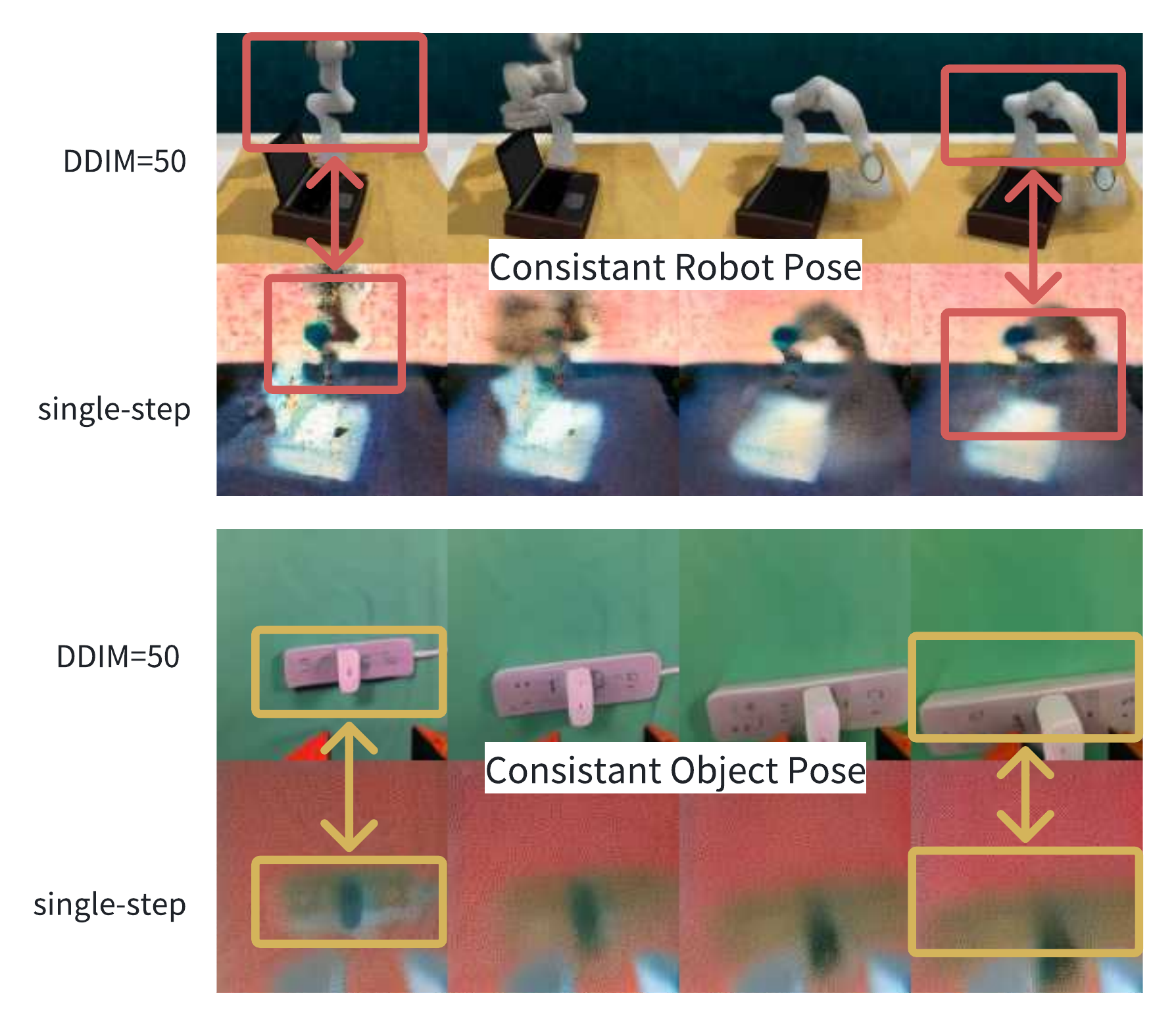}
    \caption{\textbf{Decoded frames from 50 DDIM steps vs. a single DDPM step.} Instead of generating full video frames, extracting latent features from a single diffusion step provides a semantically rich and computationally efficient representation. This single-step representation is not only fast to generate but also sufficiently informative to support downstream tasks such as VLA policy execution and failure prediction.}
    \label{fig:demo}
    \vspace{-0.5em}
\end{figure}

\begin{figure*}[t]
    \centering
    \includegraphics[width=1.0\linewidth ]{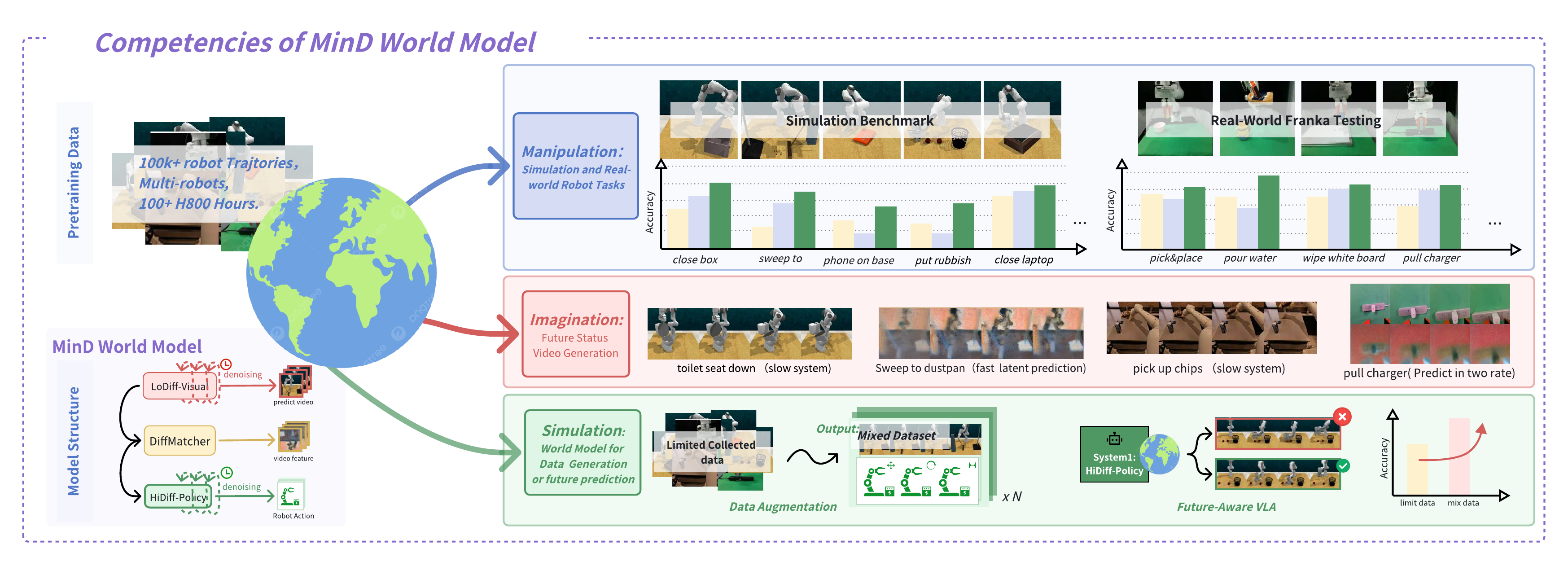}
    \caption{We present \textbf{\textit{M}}anipulate \textbf{\textit{in}} \textbf{\textit{D}}ream(\textbf{\textit{MinD}}): A video-action unified generation world model that can manipulate, imagine, and simulate. MinD integrates \textcolor{red}{\textbf{LoDiff-Visual}} for low-frequency video generation and \textcolor{DeepGreen}{\textbf{HiDiff-Policy}} for high-frequency action planning. A dynamic feature adapter, \textcolor{DeepYellow}{\textbf{DiffMatcher}}, bridges motion features between the two systems, ensuring consistency across video and action.}
    \label{fig:teaser}
    \vspace{-1.5em}
\end{figure*}

However, simply using VGMs as representation encoders fails to fully tap into their potential as world models: the inherent capability to \textbf{predict future states} through generative simulation. This oversight stems from two fundamental challenges that hinder their application in real-time robotics. First, integrating a generative process directly into policy learning is technically challenging and conceptually underdeveloped, often creating a disconnect between planning and imagination~\cite{cheang2024gr}. Second, and more pragmatically, the iterative nature of video diffusion is computationally inefficient, making naive ``imagination'' of future frames far too slow for real-world decision-making~\cite{chen2024videocrafter2, xing2023dynamicrafter, chi2024eva}.

To overcome these barriers, we challenge the assumption that a complete, fully-rendered video is necessary for control. Our key insight is that a robotic policy does not require a fully denoised future frame. Instead, the latent representation from a single denoising timestep~\cite{hu2024vpp} can provide a computationally cheap yet semantically rich signal of the predicted future, sufficient to guide a high-frequency action policy.

Based on this insight, we introduce \textbf{\textit{M}}anipulate \textbf{\textit{in}} \textbf{\textit{D}}ream (\textbf{\textit{MinD}}), a dual-system world model designed for efficient, risk-aware manipulation, as depicted in \cref{fig:teaser}. MinD operates with two asynchronous diffusion processes: a low-frequency video generator (\textbf{LoDiff-Visual}) that begins to imagine a future trajectory, and a high-frequency diffusion policy (\textbf{HiDiff-Policy}) that outputs real-time actions. The core of our framework is to condition the fast policy on the nascent, single-step predictions from the slow generator.

A critical challenge, however, is aligning these two distinct diffusion processes that operate on different schedules and temporal resolutions. To solve this, we propose the \textbf{DiffMatcher} module, which employs a novel co-training strategy. Inspired by diffusion-forcing~\cite{chen2024diffusion}, we introduce a mechanism that explicitly aligns the intermediate latent representations of both the video generator and the action policy. This forces the action model to effectively interpret the predictive signals from the video model, even when they originate from varying, early stages of the denoising process, thereby stabilizing training and ensuring coherent planning.

We validate MinD on extensive benchmarks, demonstrating its effectiveness in both simulation and the real world. Our contributions are threefold:

\textbf{(1)} We propose \textbf{MinD}, a dual-system diffusion-based world model that efficiently unlocks the predictive power of VGMs for real-time control. It unifies a low-frequency video imagination process with a high-frequency action policy.

\textbf{(2)} We introduce \textbf{DiffMatcher}, a novel module with a \textbf{diffusion-forcing} co-training strategy. It effectively aligns the asynchronous video and action diffusion processes by matching their intermediate latent representations, enabling stable and coherent policy learning.

\textbf{(3)} We demonstrate state-of-the-art performance, achieving \textbf{63\% success on RL-Bench and 60\% on real-world Franka tasks}. Critically, we show that the generated video clips allow us to identify \textbf{74\% of potential task failures in advance}, while the single-step predictive feature also demonstrates potential as an indicator for failure execution, establishing a new paradigm for building more capable, reliable, and risk-aware robotic manipulation.

\begin{figure*}[t]
    \centering
    \includegraphics[width=1.0\linewidth ]{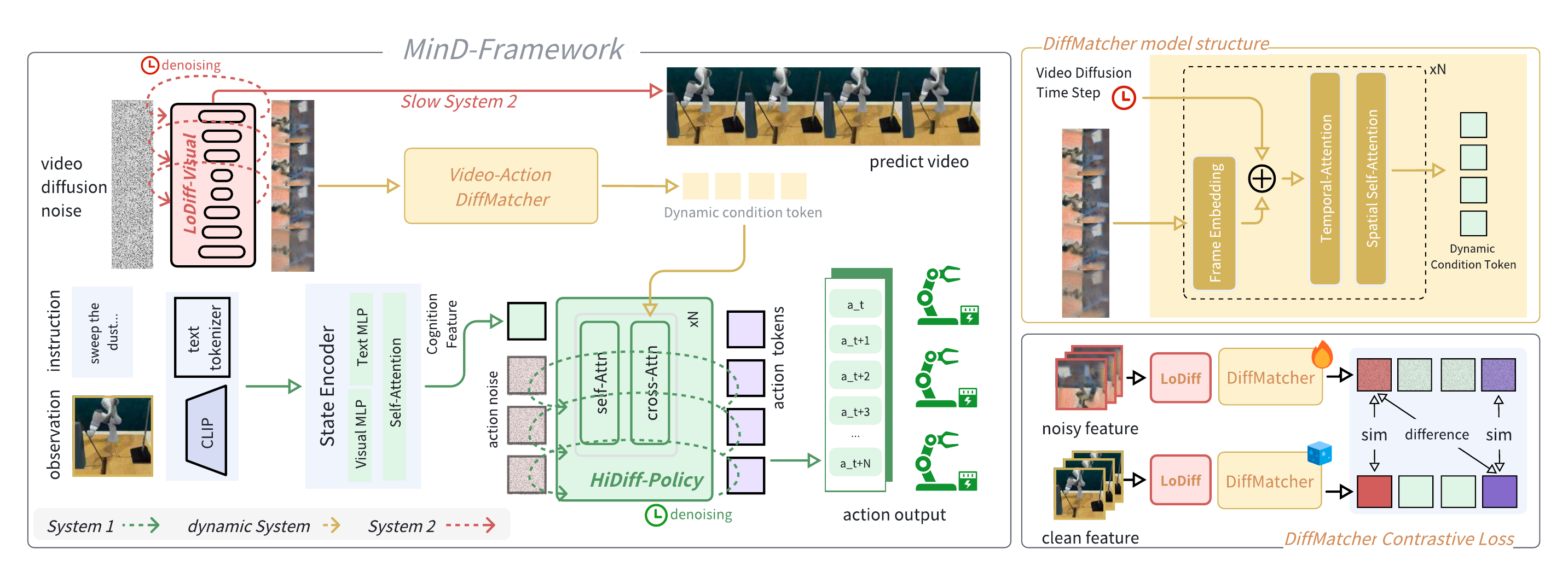}
    \caption{\textbf{MinD framework overview.} The MinD framework comprises three core components. i. Dual asynchronous diffusion models, where ``slow'' LoDiff-Visual produces future visual latents for long-latency scenes, ii. ``fast'' HiDiff-Policy outputs high-frequency actions. iii. DiffMatcher Module bridging visual and action modalities. While training, a co-training strategy employing a diffusion-forcing loss for DiffMatcher to learn mappings robust to different noise levels. We first pretrain the foundation model and the adapter on multiple robot pretraining datasets and then finetune the model on downstream tasks, including RL-Bench simulation and a real-world Franka robot.}
    \label{fig:method_main}
    \vspace{-1.5em}
\end{figure*}

\section{Related Work}

\paragraph{World Models for Embodied AI}
World models aim to provide agents with internal simulations of the environment to support prediction, planning, and decision-making~\cite{ha2018world}. Early efforts in this direction employed latent dynamics models for model-based reinforcement learning ~\cite{wu2023daydreamer}, while more recent approaches integrate vision and language to support open-world reasoning ~\cite{driess2023palm, brohan2023rt}. In the context of embodied AI, such models are increasingly required to generate coherent future states from raw visual observations, enabling long-horizon planning and closed-loop control. However, most prior work either focuses on purely reactive policies or lacks the ability to generate semantically meaningful future observations.

\paragraph{Video Generation Models for Imagination and Planning}
Video generation models (VGMs) have recently shown strong potential in synthesizing realistic and temporally coherent video sequences ~\cite{yang2023unisim, xing2023dynamicrafter,chi2024eva, li2025manipdreamer}. However, VGMs are often computationally expensive and suffer from long inference times, which limits their applicability in real-time robotics~\cite{RoboDreamer, agarwal2025cosmos,huang2025enerverse}. VPP~\cite{hu2024vpp} first introduces an efficient way to use VGM as a VLA backbone, while highly relying on multi-view for stable control. For recent hybrid frameworks like ~\cite{li2025unified, wang2025unified}, which introduce the video feature with different methods for better VLA planning, ignore the relevance of video and action, and ignore the advantages of video generation. In short, their integration with downstream control policies remains underexplored, often resulting in disjointed simulation-control pipelines.

\paragraph{Multimodal Policy Learning and Vision-Language Manipulation}
Recent advancements in VLA modeling have enabled robots to follow natural language instructions to perform manipulation tasks ~\cite{liu2025hybridvla, black2024pi_0}. Approaches such as diffusion policies ~\cite{chi2023diffusion} and transformer-based policies ~\cite{kim2024openvlaopensourcevisionlanguageactionmodel, wen2025tinyvla, gao2025adaworld} have shown promise in learning complex visuomotor behaviors. While these methods excel at conditioned action generation, they typically lack a forward model of the environment, limiting their planning capabilities. Some works attempt to bridge this gap by jointly modeling actions and visual outcomes ~\cite{yang2023unisim, li2025unified,  chi2024eva, hu2024video, cheang2024gr,wu2023unleashinggr1}, but still struggle with temporal inconsistencies and lack of alignment between imagined futures and executable behaviors.

\section{Preliminary: Diffusion Models}
\label{sec:preliminary}

MinD is built upon diffusion models~\cite{ho2020denoising}, which form the foundation for both our video generator~\cite{xing2023dynamicrafter} and our action policy~\cite{chi2023diffusion}. A diffusion model consists of a fixed forward process that gradually adds noise to data $\mathbf{x}_0$ over $T$ timesteps, and a learned reverse process, parameterized by a neural network $\epsilon_\theta$, that aims to predict and remove the noise. The model is trained by minimizing the difference between the predicted noise $\epsilon_\theta(\mathbf{x}_\tau, \tau, \mathbf{c})$ and the true noise $\epsilon$ at a given timestep $\tau$ with conditioning $\mathbf{c}$:
\begin{equation}
\mathcal{L}_{\text{diff}} = \mathbb{E}_{\mathbf{x}_0, \tau, \epsilon} \left[ \left\| \epsilon - \epsilon_\theta(\sqrt{\bar{\alpha}_\tau} \mathbf{x}_0 + \sqrt{1 - \bar{\alpha}_\tau} \epsilon,\ \tau,\ \mathbf{c}) \right\|^2 \right].
\label{eq:l_diff}
\end{equation}
Our framework utilizes two distinct diffusion models operating in a hierarchical system, which we detail next.

\section{Method: Manipulation in Dream}
\label{sec:method}
To address the challenge of leveraging generative models for efficient robotic control, we propose \textbf{MinD}, a dual-system world model. The central idea is to decouple long-latency visual imagination from high-frequency action generation, enabling the policy to be conditioned on an intermediate, computationally inexpensive future representation rather than a fully rendered video. As illustrated in Figure~\ref{fig:method_main}, the framework comprises three core components.

\subsection{Hierarchical Diffusion Framework}
\label{sec:framework}
MinD operates with two asynchronous diffusion models:

\paragraph{Low-Frequency Video Generator (LoDiff-Visual).} This model serves as the ``slow'' imagination system. It is a latent diffusion model that, given an initial observation $\mathbf{v}_0$ and a language instruction $\mathbf{l}$, predicts a sequence of future visual latents $\hat{\mathbf{v}}_{1:T'}$. It utilizes a long diffusion schedule (e.g., $T'=1000$ steps) to generate coherent and semantically rich long-horizon scenes.

\paragraph{High-Frequency Action Policy (HiDiff-Policy).} This model is the ``fast'' control system. It is a Diffusion Transformer (DiT) with fewer parameters that generates a high-frequency sequence of actions $\hat{\mathbf{a}}_{1:T''}$. It is conditioned on single-step predictive visual features from LoDiff-Visual and employs a much shorter schedule (e.g., $T''=100$ steps) to facilitate rapid action generation.

\paragraph{DiffMatcher Module.} To bridge the two asynchronous systems, we introduce DiffMatcher, a lightweight module responsible for aligning visual and action modalities. It takes a noisy visual latent $\mathbf{v}_t$ from LoDiff-Visual at a given denoising step $t$ and maps it to a compact feature vector $\mathbf{z}_t$. This feature $\mathbf{z}_t$ conditions HiDiff-Policy, effectively grounding the action plan in the predicted future state.

\subsection{Asynchronous Co-Training Strategy}
\label{sec:co_training}
A key challenge is ensuring the HiDiff-Policy can interpret the noisy, intermediate single-step representations from LoDiff-Visual. Training the models independently would fail to establish this connection. We therefore propose a co-training strategy that jointly optimizes the entire framework. The total training objective is a weighted sum of three losses:
\begin{equation}
\mathcal{L}_\text{total} = \lambda_v \mathcal{L}_\text{video} + \lambda_a \mathcal{L}_\text{action} + \lambda_{align} \mathcal{L}_\text{align}.
\end{equation}
Here, $\mathcal{L}_\text{video}$ and $\mathcal{L}_\text{action}$ are the standard diffusion losses (as in Eq.~\ref{eq:l_diff}) for the video and action, respectively.

Our primary technical contribution lies in the alignment loss, $\mathcal{L}_\text{align}$, which we term \textbf{diffusion-forcing}. During training, we simulate the conditions of inference by taking a clean ground-truth video latent $\mathbf{v}_0$ and creating a noisy version $\mathbf{v}_t$ by adding noise corresponding to a random timestep $t$. We then train the DiffMatcher, denoted by $\phi(\cdot)$, to produce a noise-invariant representation. This is achieved by minimizing the L2 distance between the matcher's output for the noisy latent and its output for the clean latent:
\begin{equation}
\mathcal{L}_\text{align} = \mathbb{E}_{\mathbf{v}_0, t} \left[ \left\| \phi(\mathbf{v}_{t}, t) - \text{sg}(\phi(\mathbf{v}_0, 0)) \right\|_2^2 \right],
\label{eq:l_align}
\end{equation}
where $\text{sg}(\cdot)$ denotes the stop-gradient operator to provide a stable target. This loss forces the DiffMatcher to learn a mapping that is robust to the noise level, enabling the policy to reliably condition on partially-denoised features.

\subsection{Inference via Single-Step Prediction}
\label{sec:inference}
The design of our framework enables a highly efficient inference pipeline. Instead of a full, multi-step denoising of the future video, we perform only a \textbf{single denoising step} on the LoDiff-Visual model. Starting from a pure noise tensor $\mathbf{v}_{T'}$, we compute the latent state for a large timestep, typically $t=T'-1$:
\begin{equation}
\mathbf{v}_{T'-1} = \text{LoDiff-Visual}(\mathbf{v}_{T'}, t=T'-1).
\end{equation}
This single-step latent $\mathbf{v}_{T'-1}$, which represents a nascent prediction of the future, is then passed through DiffMatcher:
\begin{equation}
\mathbf{z}_{T'-1} = \text{DiffMatcher}(\mathbf{v}_{T'-1}, t=T'-1).
\end{equation}
Finally, this compact feature vector conditions the HiDiff to generate the entire action sequence in a single forward pass:
\begin{equation}
\hat{\mathbf{a}}_{1:T''} = \text{HiDiff-Policy}(\mathbf{z}_{T'-1}).
\end{equation}
This process allows MinD to leverage the powerful predictive capabilities of the video model at a computational cost suitable for real-time robotic control.

\subsection{Implicit Risk Assessment via Latent Feature Analysis}
\label{sec:risk}
We hypothesize that the latent feature $\mathbf{v}_{t}$ produced by the world model encodes not only future predictions but also uncertainty. A well-structured latent distribution often indicates confident predictions, while anomalies in the latent space suggest ambiguity or risk. PCA analysis reveals meaningful structure in $\mathbf{v}_{t}$, supporting its potential for uncertainty modeling. Additionally, we find that visual failures in generated rollouts often correlate with downstream errors. This highlights the potential of using world models as implicit distribution predictors, enabling lightweight risk assessment without extra supervision.
\definecolor{lightgray}{gray}{.9}
\definecolor{lightblue}{RGB}{230,240,255}
\definecolor{lightgreen}{RGB}{230,255,230}
\definecolor{lightyellow}{RGB}{255,255,230}
\definecolor{lightred}{RGB}{255,230,230}

\definecolor{lightlightgray}{gray}{.95}
\definecolor{lightlightblue}{RGB}{240,245,255}
\definecolor{lightlightgreen}{RGB}{240,255,240}
\definecolor{lightlightyellow}{RGB}{255,255,240}
\definecolor{lightlightred}{RGB}{255,240,240}

\definecolor{lightlightlightgray}{gray}{.99}
\definecolor{lightlightlightblue}{RGB}{247,250,255}
\definecolor{lightlightlightgreen}{RGB}{247,255,247}
\definecolor{lightlightlightyellow}{RGB}{255,255,247}
\definecolor{lightlightlightred}{RGB}{255,247,247}

\begin{table*}[t]
\captionsetup{position=top} 
\caption{We compare the performance of MinD with existing VLA models across 7 tasks in RLBench settings. For the model using Mamba or LLM as the backbone, we colored it with a green background. We use a yellow background for the VLA models with a video generation backbone, and a red background for our method.} 
\label{tab:rlbench}
\resizebox{\linewidth}{!}{%
\setlength{\tabcolsep}{2mm}{
\begin{tabular}{ccccccc}
\toprule\toprule
\multirow{2}{*}{\textbf{Methods}} & Pretrain & \multirow{2}{*}{Action Head} & \multirow{2}{*}{Backbone} & \cellcolor{lightgray}Phone & \cellcolor{lightgray}Toilet & \cellcolor{lightgray}Close \\
 & modality & & & \cellcolor{lightgray}on base & \cellcolor{lightgray}seat down & \cellcolor{lightgray}box \\
\midrule\midrule
\rowcolor{lightlightlightgreen}
\multicolumn{1}{c}{\cellcolor{lightgreen}OpenVLA~\cite{kim2024openvlaopensourcevisionlanguageactionmodel}~\textcolor{gray}{(CoRL'24)}} & V+A & MLP & LLaMA2-7B & 20\% & 76\% & 72\% \\
\rowcolor{lightlightlightgreen}
\multicolumn{1}{c}{\cellcolor{lightgreen}CogAct-S~\cite{li2024cogactfoundationalvisionlanguageactionmodel}~\textcolor{gray}{(Arxiv'24)}} & V+A & Diffusion & LLaMA2-7B & 52\% & 44\% & 84\% \\
\rowcolor{lightlightlightgreen}
\multicolumn{1}{c}{\cellcolor{lightgreen}CogAct-L~\cite{li2024cogactfoundationalvisionlanguageactionmodel}~\textcolor{gray}{(Arxiv'24)}} & V+A & Diffusion & LLaMA2-7B & 56\% & 100\% & 64\% \\
\rowcolor{lightlightlightgreen}
\multicolumn{1}{c}{\cellcolor{lightgreen}RoboMamba~\cite{liu2024robomambaefficientvisionlanguageactionmodel}~\textcolor{gray}{(NeruIPS'24)}} & V+A & MLP & Mamba-2.8B  & 44\% & 64\% & 60\% \\
\rowcolor{lightlightlightgreen}
\multicolumn{1}{c}{\cellcolor{lightgreen}Pi-0~\cite{black2024pi0visionlanguageactionflowmodel}~\textcolor{gray}{(Arxiv'24)}} & V+A & Flow-Matching & LLaMA2-7B & 32\% & 96\% & 88\% \\
\rowcolor{lightlightlightyellow}
\rowcolor{lightlightlightyellow}
\multicolumn{1}{c}{\cellcolor{lightyellow}RoboDreamer~\cite{RoboDreamer}~\textcolor{gray}{(ICML'24)}} & V & MLP & Video Diffusion-400M  & 64\% & 32\% & 88\%  \\
\rowcolor{lightlightlightyellow}
\multicolumn{1}{c}{\cellcolor{lightyellow}RoboDreamer-faster~\cite{RoboDreamer}~\textcolor{gray}{(ICML'24)}}& V & MLP & Video Diffusion-400M & 16\% & 4\% & 64\%\\
\rowcolor{lightlightlightyellow}
\multicolumn{1}{c}{\cellcolor{lightyellow}VPP~\cite{hu2024vpp}~\textcolor{gray}{(ICML'25)}}& V & Diffusion & Video Diffusion-800M & 44\% & 16\% & 92\%\\
\rowcolor{lightlightlightred}
\multicolumn{1}{c}{\cellcolor{lightred}MinD-S} & V & Diffusion & UNet-1.5B &  72\% & 48\% & 84\% \\
\rowcolor{lightlightlightred}
\multicolumn{1}{c}{\cellcolor{lightred}MinD-B} & V & Diffusion & UNet-1.5B &  60\% & 32\% & 100\% \\
\midrule
\multirow{2}{*}{\textbf{Methods}} & \cellcolor{lightgray}Put & \cellcolor{lightgray}Take umbrella & \cellcolor{lightgray}Close  & \cellcolor{lightgray}Sweep &  \multirow{2}{*}{Mean Acc.\% $\uparrow$} &  \multirow{2}{*}{FPS (Hz) $\uparrow$} \\
& \cellcolor{lightgray}rubbish in bin & \cellcolor{lightgray}out of stand & \cellcolor{lightgray}laptop lid & \cellcolor{lightgray}to dustpan &  & \\
\midrule
\rowcolor{lightlightlightgreen}
\multicolumn{1}{c}{\cellcolor{lightgreen}OpenVLA~\cite{kim2024openvlaopensourcevisionlanguageactionmodel}~\textcolor{gray}{(CoRL'24)}} & 8\% & 28\% & 64\% & 68\%  & 48.0\% & 6.3 \\
\rowcolor{lightlightlightgreen}
\multicolumn{1}{c}{\cellcolor{lightgreen}CogAct-S~\cite{li2024cogactfoundationalvisionlanguageactionmodel}~\textcolor{gray}{(Arxiv'24)}} & 44\% & 40\% & 60\% & 64\% &  55.4\% & 9.6 \\
\rowcolor{lightlightlightgreen}
\multicolumn{1}{c}{\cellcolor{lightgreen}CogAct-L~\cite{li2024cogactfoundationalvisionlanguageactionmodel}~\textcolor{gray}{(Arxiv'24)}} & 60\% & 32\% & 76\% & 44\% & \textcolor{blue}{61.7\%} & 8.6 \\
\rowcolor{lightlightlightgreen}
\multicolumn{1}{c}{\cellcolor{lightgreen}RoboMamba~\cite{liu2024robomambaefficientvisionlanguageactionmodel}~\textcolor{gray}{(NeruIPS'24)}} & 36\% & 32\% & 52\% & 32\% & 45.7\% & - \\
\rowcolor{lightlightlightgreen}
\multicolumn{1}{c}{\cellcolor{lightgreen}Pi-0~\cite{black2024pi0visionlanguageactionflowmodel}~\textcolor{gray}{(Arxiv'24)}} & 48\% & 32\% & 80\% & 28\% & 57.7\% & 13.8 \\
\rowcolor{lightlightlightyellow}
\multicolumn{1}{c}{\cellcolor{lightyellow}RoboDreamer~\cite{RoboDreamer}~\textcolor{gray}{(ICML'24)}} & 32\% & 40\% & 68\% & 76\% & 50.3\% & 0.7 \\
\rowcolor{lightlightlightyellow}
\rowcolor{lightlightlightyellow}
\multicolumn{1}{c}{\cellcolor{lightyellow}RoboDreamer-faster~\cite{RoboDreamer}~\textcolor{gray}{(ICML'24)}}  & 24\% & 20\% & 60\% & 64\% & 37.4\% & 1.1 \\
\multicolumn{1}{c}{\cellcolor{lightyellow}VPP~\cite{hu2024vpp}~\textcolor{gray}{(ICML'25)}}  & 40\% & 12\% & 84\% & 80\% & 53.0\% & 1.1 \\
\rowcolor{lightlightlightred}
\multicolumn{1}{c}{\cellcolor{lightred}MinD-S}  & 32\% & 24\% & 60\% & 84\% & 58.0\%\textcolor{gray}{($- 2.4\%$)} & 11.3 \\
\rowcolor{lightlightlightred}
\multicolumn{1}{c}{\cellcolor{lightred}MinD-B} & 52\% & 32\% & 68\% & 96\% & \textcolor{red}{\textit{\textbf{63.0\%}}}\textcolor{gray}{($\pm 1.5\%$)} & 10.2 \\
\bottomrule\bottomrule
\end{tabular}%
}}
\vspace{-1em}
\end{table*}


\begin{figure*}[ht]
    \centering
    \includegraphics[width=1.\linewidth ]{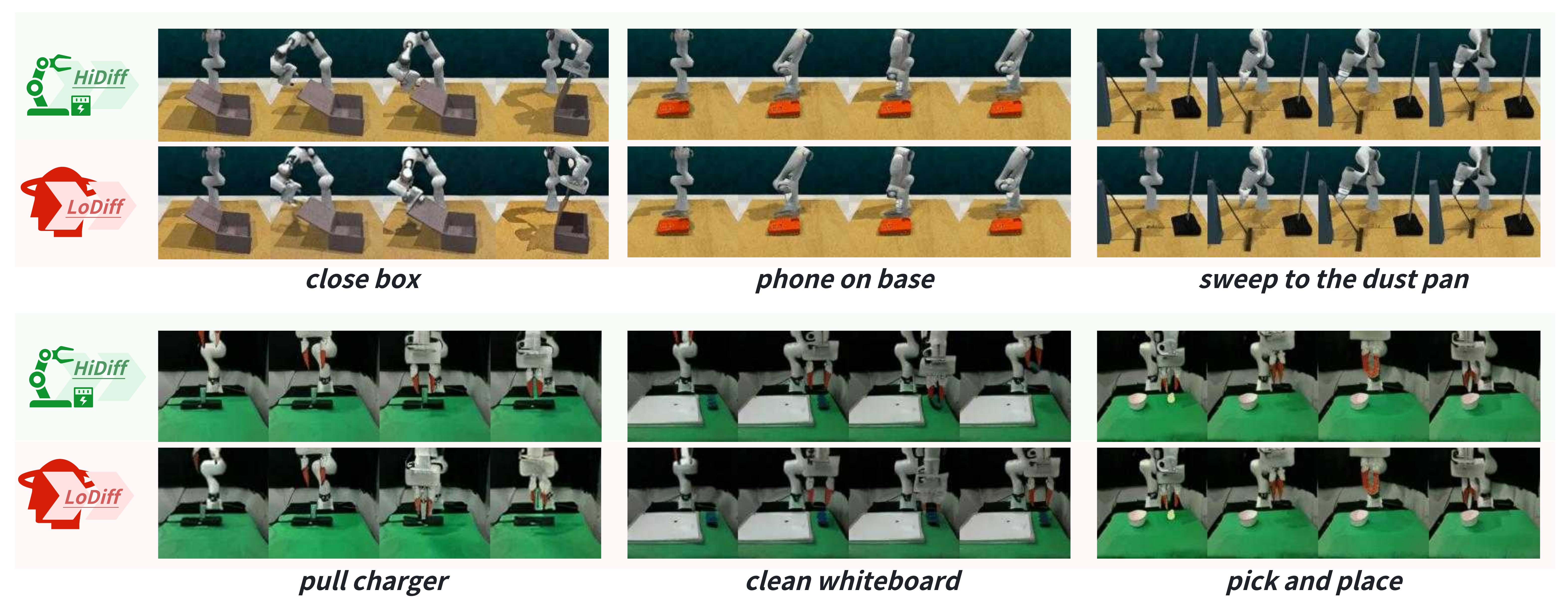}
    \caption{This figure showcases the consistency between the future imagined by our LoDiff video generator (bottom rows) and the final trajectory executed by the HiDiff policy (top rows). This high-fidelity alignment across both RLBench simulation and real-world Franka tasks validates MinD's capability as an effective world model.}
    \vspace{-1em}
    \label{fig:exp_main}
\end{figure*}

\section{Experiments}
\label{sec:exp}
In this section, we evaluate the performance of our proposed MinD framework through extensive experiments in both real-world and simulation environments. These experiments focus on manipulation tasks, comparing MinD with existing VLA models across multiple benchmarks.

\subsection{Model Implementation Details}

Our model builds upon the Dynamicrafter framework~\cite{xing2023dynamicrafter}, preserving its visual encoder and spatial-temporal UNet backbone as the pretrained visual module. For robot task training, we follow OpenVLA~\cite{kim2024openvlaopensourcevisionlanguageactionmodel}, using CLIP~\cite{radford2021learning} features and tokenizer for efficient state representation.

We implement two HiDiff-Policy variants (\textit{Small} and \textit{Base}), matching CogACT~\cite{li2024cogactfoundationalvisionlanguageactionmodel} model sizes with added cross-attention layers. Data preprocessing follows Dynamicrafter, including 128×128 image resizing and RLDS~\cite{ramos2021rlds} format for actions.

\paragraph{Pretraining.} We apply feature alignment pretraining using RT-1~\cite{brohan2023rt}, Robomind~\cite{wu2024robomind}, and parts of OXE~\cite{o2024oxe}. Training uses PyTorch 2.5.1 with CUDA 12.1, on 4×A100 GPUs, batch size 16, AdamW~\cite{adamw}, learning rate $2\mathrm{e}{-5}$ with cosine decay and 2k warmup steps. Models are trained for 50k steps and then fine-tuned on downstream tasks (details in Appendix).

\subsection{Manipulation in Silumation}
\label{exp_rlbench}

\paragraph{Simulation Benchmark Setting}
We evaluate our method on RL-Bench~\cite{james2020rlbench} using CoppeliaSim, with a Franka Panda robot and multiview cameras. We use only front-view images as input. Seven diverse tasks are selected, and data is constructed using predefined waypoints and key-frame downsampling~\cite{goyal2023rvt,zhang2025mole}. In total, we collect 1000 trajectories: 100 per task and 300 additional from randomly sampled tasks.

\paragraph{Fine-tuning and Evaluation}
The pretrained checkpoint is fine-tuned on the constructed dataset using the Adam optimizer~\cite{adam} with a learning rate of $2\mathrm{e}{-5}$ for 10k steps. Images are resized to $128 \times 128$, with standard augmentations (random crop, flip, color jitter). Performance is measured by task success rate over 25 trials, averaged across three runs. We report results for two MinD variants with different HiDiff-Policy sizes; detailed model specs are provided in the Appendix.

\paragraph{Baselines} 
We compare our model with several VLA baselines, including OpenVLA~\cite{kim2024openvlaopensourcevisionlanguageactionmodel}, Pi-0~\cite{black2024pi0visionlanguageactionflowmodel}, and CogACT~\cite{li2024cogactfoundationalvisionlanguageactionmodel}. Besides, we also conducted comparisons between our model and VGM-based world models such as RoboDreamer~\cite{RoboDreamer} and VPP~\cite{hu2024video}. These baselines employ different backbone architectures and training paradigms, such as transformer-based models or convolutional neural networks. We use the same RL-Bench tasks and evaluation metrics for all methods under the same setting as described above.

\paragraph{Inference Speed} 
We measure the average step prediction time on an RTX-4090 GPU. Notably, MinD-S achieves an inference speed of \textbf{11.3 FPS}, showcasing its superior efficiency. In contrast, VGM-based world models like RoboDreamer and VPP are limited in speed, running at approximately 1 FPS and 6.5 FPS, respectively. While reducing its video generation steps (e.g., decreasing DDIM steps from 100 to 50) slightly increases FPS, it significantly degrades the success rate. This highlights the effectiveness of our proposed DiffMatcher and dual-schedule design.

\paragraph{Quantitative Results} 
~\cref{tab:rlbench} shows that our models, MinD-S and MinD-B, achieve SOTA performance on RLBench tasks among three different types of VLA models. MinD-B achieves the highest mean accuracy of \textbf{63.0\%}, while MinD-S follows with \textbf{58.0\%}, outperforming all baseline methods. Notably, our models excel in tasks requiring complex temporal reasoning, such as "Sweep to Dustpan" (\textbf{96\%}) and "Close Laptop Lid" (\textbf{68\%}). These results highlight that video generation models can serve as a strong foundational backbone for comprehensive VLA tasks.

\subsection{Manipulation in Real-world Robotics}

\paragraph{Real-World Franka Robot Setup} 
We use a Franka Research 3 robot for the experiment. The robot is set up with a front-view and a wrist-view camera. For training and testing, we use three distinct types of tasks: \textit{1) a pick-and-place task (e.g., placing a cube into a bowl or a croissant into a basket), 2) unplugging the charger, and 3) wiping the whiteboard}. For each task, we collect 100 human demonstration trajectories via teleoperation using a SpaceMouse.

\paragraph{Training and Evaluation Details} 
We fine-tuned the pretrained checkpoint on the self-collected fine-tuning dataset. The fine-tuning process and parameter setups follow the RL-Bench experiments as described in previous sections. For each single task, we conducted over 20 rollouts using the latest checkpoints across varied tabletop positions. Both training and evaluation processes are conducted in the same robotic setup to ensure consistency between data collection and testing conditions.

\paragraph{Quantitative Results} 
As demonstrated in the table, MinD achieves success rates of \textbf{68.75\%} and \textbf{72.5\%} with wrist and front-view cameras, respectively. This performance consistently surpasses that of VLA models, represented by OpenVLA, and VGM-based world models, such as VPP. Notably, while VPP also shows competitive results on the long-horizon \textit{wipe blackboard} task, our model attains superior performance and operates at a higher control frequency. This is attributed to our DiffMatcher design, which preserves the characteristics of both fast and slow systems, enabling a tighter coupling between prediction and execution.

\begin{table}[tb]
    \centering
    \caption{Real-world Franka robot execution success rate comparison across three types of tasks: \textit{pick and place, unplug the charger, and wipe the whiteboard}. Results are shown for OpenVLA, Video Prediction Policy(VPP), and MinD. Under limited training data, MinD demonstrates competitive performance.}
    \label{tab:franka}
    \resizebox{0.99\linewidth}{!}{
    \setlength{\tabcolsep}{2mm}
    \begin{tabular}{cccccc}
    \toprule\toprule
    \multirow{2}{*}{\textbf{Methods}} & \multicolumn{2}{c}{\textbf{Pick \& Place}} & \textbf{Unplug} & \textbf{Wipe} & \multirow{2}{*}{\textbf{Average}} \\
    \cmidrule(lr){2-3} 
    & \textbf{Cube} & \textbf{Croissant} & \textbf{Charger} & \textbf{Whiteboard} & \\
    \midrule\midrule
    \rowcolor{lightlightlightgreen}
    \cellcolor{lightgreen}OpenVLA & 40\% & 55\% & 25\% & 30\% & 37.50\% \\
    \rowcolor{lightlightlightgreen}
    \cellcolor{lightgreen}VPP & 55\% & 65\% & 40\% & 60\% & 52.50\% \\
    \rowcolor{lightlightlightred}
    \cellcolor{lightred}MinD-FrontView & 60\% & 80\% & 50\% & 85\% & \textcolor{red}{\textit{\textbf{68.75\%}}} \\
    \rowcolor{lightlightlightred}
    \cellcolor{lightred}MinD-WristView & 75\% & 85\% & 65\% & 65\% & \textcolor{red}{\textit{\textbf{72.50\%}}} \\
    \bottomrule\bottomrule
    \end{tabular}%
    }
\end{table}
\subsection{Ablation Study}
The results of the ablation study are presented in ~\cref{tab:ablation}. Each row corresponds to a different configuration of MinD, with variations in modalities (action-only, video-only, or both) and trainable modules. The evaluation metrics include the Fréchet Video Distance (FVD), which measures the quality of generated videos, and the success rate (SR), which quantifies task execution performance. Quality comparison can be found in Appendix A.

1. \textbf{Large scale video data Pretraining (LDP)}: The inclusion of large-scale data pretraining significantly improves both FVD and SR. For instance, the configuration with A+V, LDP, and all loss terms achieves the highest SR (64.0\%) while maintaining competitive FVD (378.3).

2. \textbf{Diffusion Modules}: The combination of LoDiff, DiffMatcher, and HiDiff plays a crucial role in improving task execution success. Configurations without these modules (e.g., rows with ``-'' for LoDiff or DiffMatcher) show a marked drop in performance.

3. \textbf{Loss Functions}: The ablation of specific loss terms also impacts performance. For example, removing $\mathcal{L}_\text{video}$ and $\mathcal{L}_\text{sim}$ (last row) leads to a lower SR (58.3\%) and a significantly higher FVD (596.5), indicating the importance of these loss terms for aligning video generation with task requirements.

4. \textbf{Preservation of VLA and World Model}: The inclusion of both $\mathcal{L}_\text{video}$ and $\mathcal{L}_\text{sim}$ ensures that the system retains the functionalities of manipulation(SR 63.4\%) and the video prediction(FVD 307.1). Ablating these losses, as seen in the last row, leads to a drop in SR (58.3\%) and a sharp increase in FVD (596.5), indicating degraded performance in both video generation and task execution.

Overall, the results demonstrate that incorporating both modalities, pretraining, and a full set of loss terms is critical to achieving an optimal success rate and generation quality. These findings provide insights into the design of future multimodal frameworks for robotic learning.

\begin{table}
\centering
\caption{Ablation study of MinD across different configurations of modalities (A: action, V: video) and trainable modules. SE denotes the state encoder, and LDP represents large-scale data pretraining. We evaluate each configuration based on video generation quality (FVD~\cite{unterthiner2019fvd}) and success rate (SR) in task execution. The results highlight the impact of key components such as LDP, diffusion modules (LoDiff, DiffMatcher, HiDiff), and loss functions ($\mathcal{L}_\text{video}$, $\mathcal{L}_\text{sim}$, $\mathcal{L}_\text{action}$) on performance.}
\label{tab:ablation}
\setlength{\tabcolsep}{1.5mm} 
\renewcommand{\arraystretch}{1.2} 
\resizebox{\linewidth}{!}{%
\begin{tabular}{cccccccccccc}
\toprule \toprule
\multirow{2}{*}{modality} & \multicolumn{5}{c}{trainable module} & \multicolumn{3}{c}{Loss} & \multirow{2}{*}{FVD $\downarrow$} & \multirow{2}{*}{SR $\uparrow$} \\
\cmidrule(lr){3-6} \cmidrule(lr){7-9}
 & LDP & LoDiff & DiffMatcher & HiDiff & SE & $\mathcal{L}_\text{video}$ & $\mathcal{L}_\text{sim}$ & $\mathcal{L}_\text{action}$ & & \\
\midrule \midrule
\rowcolor{lightlightlightgreen}
A & - & - & \checkmark & \checkmark & - & - & - & \checkmark & -- & 44.0\% \\
\rowcolor{lightlightlightgreen}
V & \checkmark & \checkmark & - & - & - & \checkmark & - & - &  \textbf{235.5} & - \\
\rowcolor{lightlightlightgreen}
V & - & \checkmark & - & - & - & \checkmark & - & - & 352.6 & - \\
\midrule 
\rowcolor{lightlightlightgreen}
A+V&  & \checkmark & \checkmark &\checkmark & \checkmark & \checkmark & \checkmark & \checkmark & 393.3 & 55.4\% \\
\rowcolor{lightlightlightgreen}
A+V& \checkmark & \checkmark & \checkmark &\checkmark & \checkmark & - & \checkmark & \checkmark & 378.3 & \textbf{64.0\%} \\
\rowcolor{lightlightlightgreen}
A+V& \checkmark & \checkmark & \checkmark &\checkmark & \checkmark & - & - & \checkmark & 596.5 & 58.3\% \\
\rowcolor{lightlightlightred}
A+V& \checkmark & \checkmark & \checkmark &\checkmark & \checkmark & \checkmark & \checkmark & \checkmark & 307.1 & \textcolor{blue}{63.4\%} \\
\bottomrule
\end{tabular}
}
\end{table}

\begin{figure*}[!t]
    \centering
    \includegraphics[width=\linewidth ]{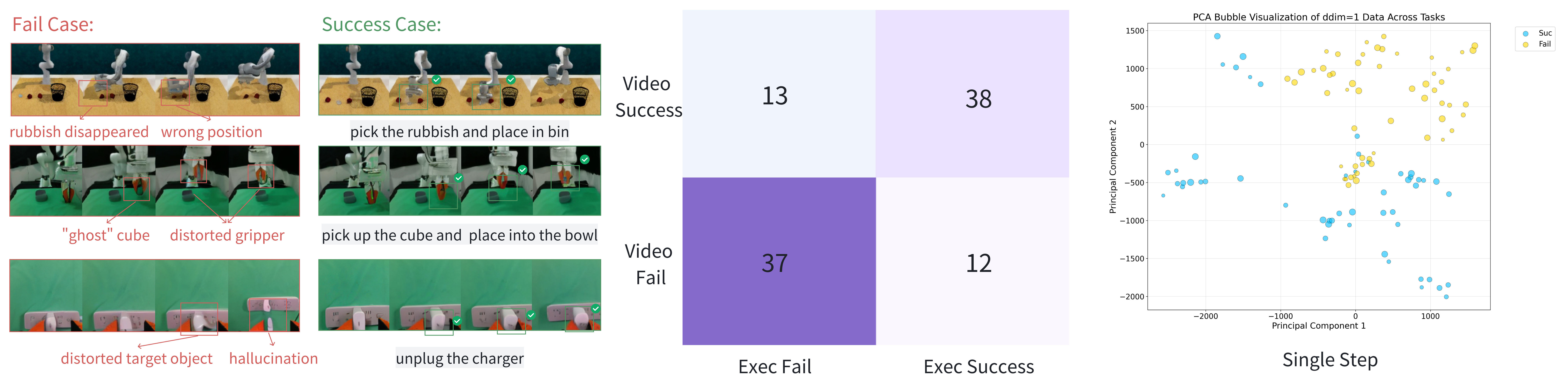}
    \caption{Evaluation of video generation predictions. The left panel visualizes failing cases (left) with misaligned generated video clips and corresponding successful cases (right) with accurate predictions. The middle panel shows the confusion matrix of our human evaluation. The right panel showcases the PCA result of the single-step predictive visual feature.}
    \label{fig:exp_fail}
\end{figure*}

\section{Video Generation Enables Explainable Risk-Aware VLA}

World-model-based VLA aims to improve interpretability and decision-making in robotics. In this section, we investigate whether VGMs can enhance trustworthiness by predicting task outcomes and identifying potential risks. Using RLBench and a Franka Panda robot across 15+ task executions, we conducted human evaluation on 100 cases. The results demonstrate that VGMs can provide actionable insights for safer control.

\paragraph{Dual-System Video Feature for Risk Assessment:}
As ~\cref{fig:exp_fail}, VGMs can align predicted videos with real-world outcomes. We first analyze the LoDiff-Visual generation results by a human. Among 50 successful executions, 38 are correctly identified (76\% true positive rate). For 50 failed cases, 37 are accurately marked (74\% true negative rate). These results suggest that VGMs can serve as effective proxies for risk estimation by anticipating execution-level failures through visual inconsistencies.

Beyond visual inspection and human evaluation, we further analyze the latent features via PCA, a single-step feature that is used for HiDiff-Policy. We observe clear distributional differences between successful and failed predictions, indicating that the latent representations encode uncertainty relevant to task outcomes. This supports the potential of modeling VLA with a dual-system perspective: fast, high-level alignment from textual goals to actions, and slow, generative simulation for dynamic risk estimation.

\paragraph{Failure Case Study.} In the left panel of~\cref{fig:exp_fail}, we present a comparison of video clips generated during successful versus failed executions under three identical scenarios. The left column displays clips from failed executions, where we can clearly observe artifacts such as distorted target objects, erroneous rotations, and hallucinations (e.g., disappearing items and "ghost" objects that suddenly appear). These flawed video predictions often lead to the failure of the task. In contrast, the videos in the right column do not exhibit these issues. These examples highlight both the predictive power of VGMs and the current limitations in handling fine-grained dynamics. Additional qualitative results are shown in~\cref{fig:exp_main}, where higher-quality predictions often correlate with successful executions.

\paragraph{Toward Safer and More Trustworthy VLA.}
VGM offers a promising path toward explainable and risk-aware VLA systems by simulating future observations. Our MinD framework demonstrates that a dual-system design—combining a slow generative pathway and a fast reactive policy—can help anticipate failures at multiple levels. We observe that poor-quality video predictions from the LoDiff often correlate with downstream task failure, while latent single-step video features passed to the HiDiff show potential as diagnostic signals. These findings highlight a new direction for building safer and more trustworthy embodied agents through cross-modal generative modeling. 
\section{Conclusion}

We present \textbf{Manipulate in Dream (MinD)}, a dual-system diffusion-based world model for vision-language-action tasks. MinD addresses key challenges of slow video generation and poor video-action consistency by combining a low-frequency visual generator with a high-frequency diffusion policy, connected through the proposed DiffMatcher module.
MinD enables real-time control and risk-aware planning, achieving strong performance on both simulation and real-world benchmarks. Beyond execution, it can anticipate failures through latent feature analysis, demonstrating the potential of video generation models as predictive and interpretable world models for robotics.
\paragraph{Limitation}  
\label{sec:limitation}
Despite its advancements, MinD is still constrained by the availability and diversity of robot training data, which limits its ability to generalize from arbitrary video inputs to robotic actions. The reliance on domain-specific datasets and task configurations hinders the model's scalability to more general-purpose video-to-robot applications. Future work should explore methods to enhance data diversity, incorporate large-scale multimodal pretraining, and improve generalization across unseen tasks and environments. These efforts are essential for building an open-world world model that can generalize beyond robotic datasets to broader, real-world scenarios.

\clearpage
\bibliography{aaai2026}

\clearpage


\definecolor{lightgray}{gray}{.9}
\definecolor{lightblue}{RGB}{230,240,255}
\definecolor{lightgreen}{RGB}{230,255,230}
\definecolor{lightyellow}{RGB}{255,255,230}
\definecolor{lightred}{RGB}{255,230,230}

\definecolor{lightlightgray}{gray}{.95}
\definecolor{lightlightblue}{RGB}{240,245,255}
\definecolor{lightlightgreen}{RGB}{240,255,240}
\definecolor{lightlightyellow}{RGB}{255,255,240}
\definecolor{lightlightred}{RGB}{255,240,240}

\definecolor{lightlightlightgray}{gray}{.99}
\definecolor{lightlightlightblue}{RGB}{247,250,255}
\definecolor{lightlightlightgreen}{RGB}{247,255,247}
\definecolor{lightlightlightyellow}{RGB}{255,255,247}
\definecolor{lightlightlightred}{RGB}{255,247,247}

\section{Appendix A: Extensive Experiments}
\label{sec:app_moreexp}

\begin{figure*}[h]
    \centering
    \includegraphics[width=1.\linewidth ]{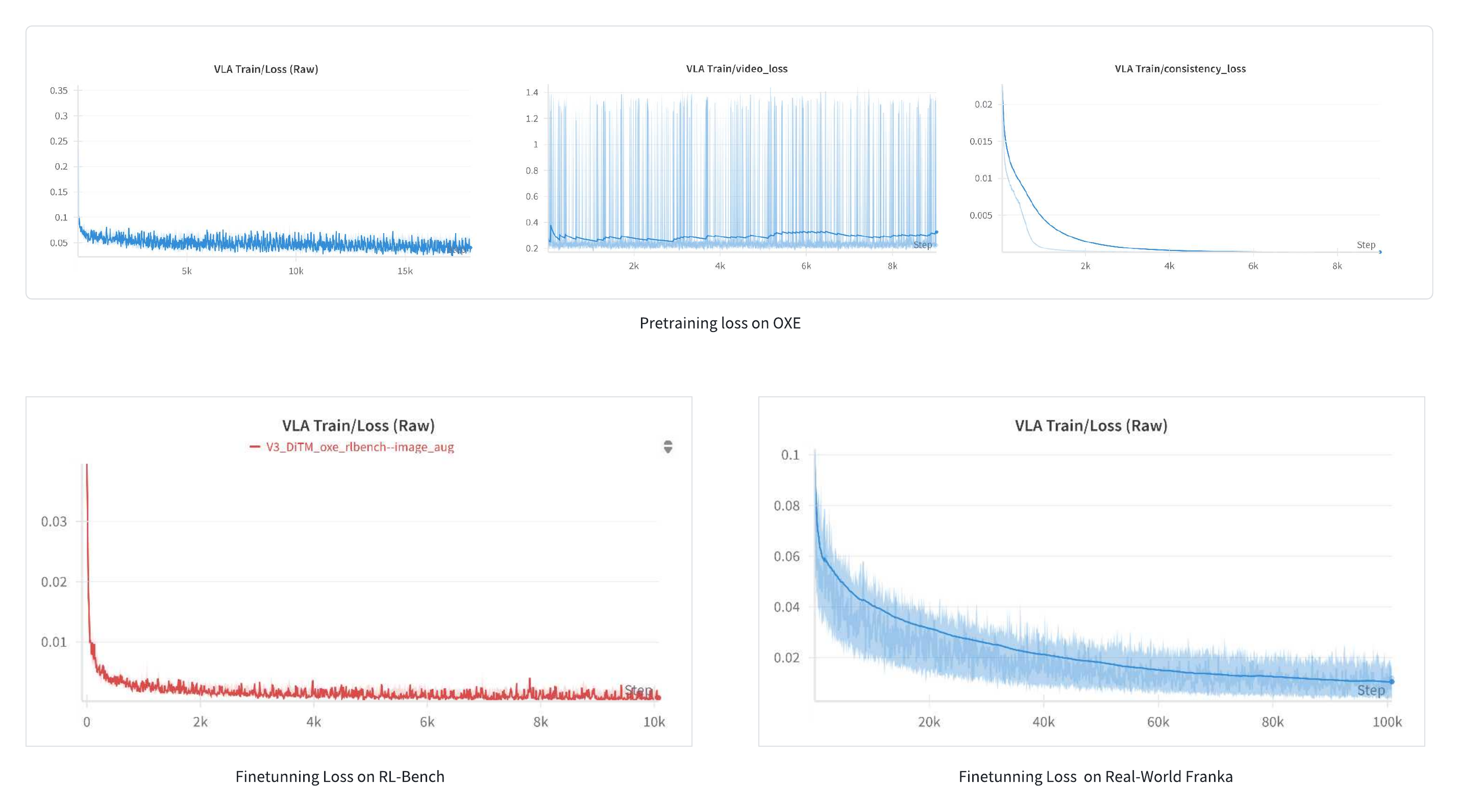}
    \caption{Loss curve of pre-training and fine-tuning.  }
    \vspace{-1em}
    \label{fig:app_loss}
\end{figure*}

\begin{figure*}[h]
    \centering
    \includegraphics[width=1.\linewidth ]{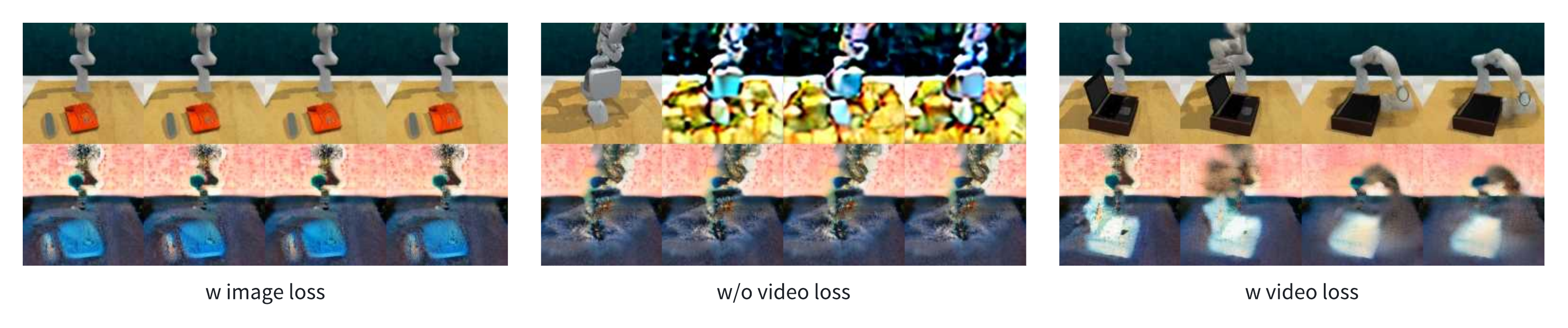}
    \caption{Qualitative comparison of video prediction ability with the real-execution results. }
    \vspace{-1em}
    \label{fig:visloss}
\end{figure*}

\begin{figure*}[h]
    \centering
    \includegraphics[width=1.\linewidth ]{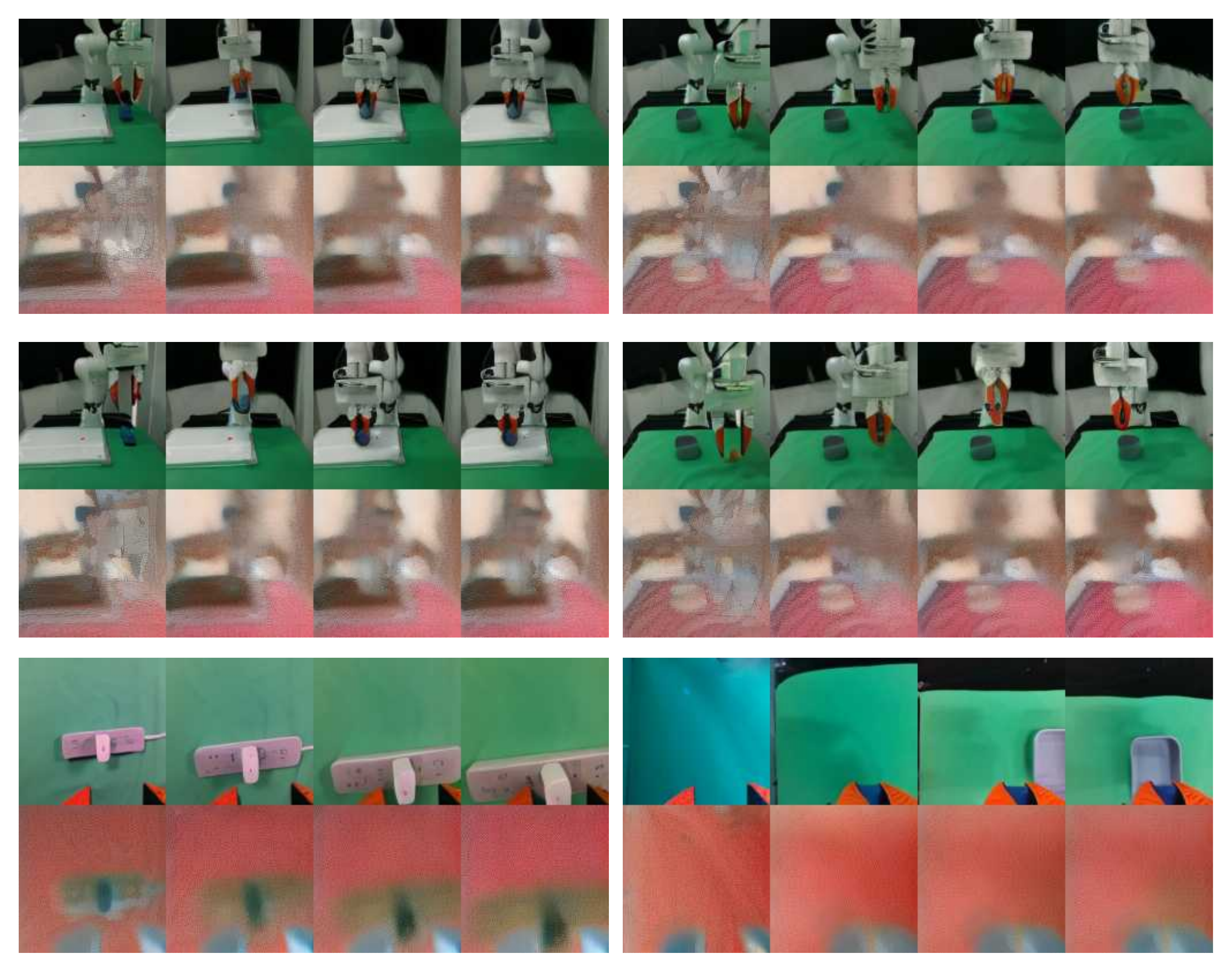}
    \caption{More visualization results of the video generation in feature level or DDIM=100. The first two rows are the front view, last row is the wrist view. We noticed that the wrist view could significantly reduce the strict pre-setting of the camera pose and position initialization. }
    \vspace{-1em}
    \label{fig:app_vgen}
\end{figure*}


\subsection{More visualization results}
\label{sec:app_vis}
While wrist view introduces challenges such as rapid scene shifts, it reduces the need for complex setups like external camera calibration. It naturally provides precise relative position cues between the gripper and the object, enabling better generalization in generation models. This leads to tighter alignment between video generation and control, making the output more actionable and transferable to real-world manipulation.

Moreover, we shows the feature of the video generation model under different loss settings~\cref{fig:visloss}.
If we do not preprocess the RLDS dataset into video format, and directly use the image loss, the video model would become a fixed frame model without temporal information. While the task successful rate still remains better than ordinary diffusion policy without temporal cross-attention insertion. If, without any video loss, the video feature would become noisy, as shown in the second figure. Using the video loss would help the MinD to retain its video generation ability.

\subsection{Failing case Prediction}
\begin{figure*}[h]
    \centering
    \includegraphics[width=1.\linewidth ]{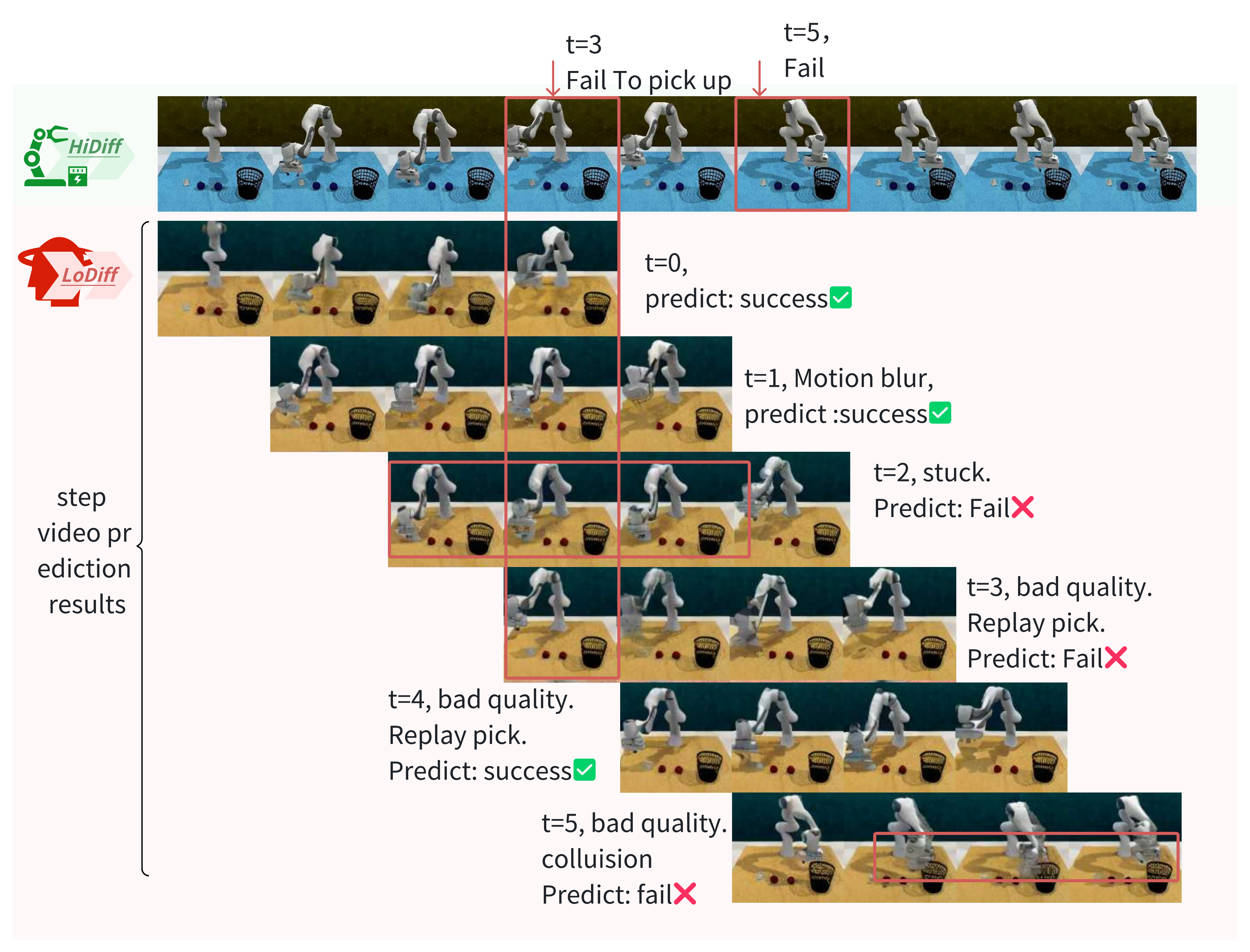}
    \caption{Qualitative comparison of video prediction ability with the real-execution results. }
    \vspace{-1em}
    \label{fig:app_fail}
\end{figure*}
 ~\cref{fig:app_fail}illustrates a qualitative comparison between real-world robot execution results (HiDiff) and predicted video rollouts (LoDiff) for manipulation tasks. While HiDiff shows the actual outcomes—such as missed grasps, obstruction, or collisions—LoDiff predicts these failures in advance.

At t = 2, for example, the robot in HiDiff fails to pick up the object. LoDiff, despite being a forward video prediction model, successfully forecasts this failure by showing the gripper getting stuck or misaligned even before the actual failure occurs. Similarly, LoDiff captures issues such as motion blur, replayed picking actions, and eventual collisions in later steps (t = 3 to t = 5), aligning with the error patterns observed in HiDiff.

This demonstrates that LoDiff can serve as an early warning signal, allowing the system to anticipate and potentially avoid failure before real-world execution, by detecting subtle cues like object shifts, bad grasps, or trajectory deviations.

\subsection{Further Discussion of LoDiff and HiDiff Video Features}
\begin{figure*}[h]
    \centering
    \includegraphics[width=1.\linewidth ]{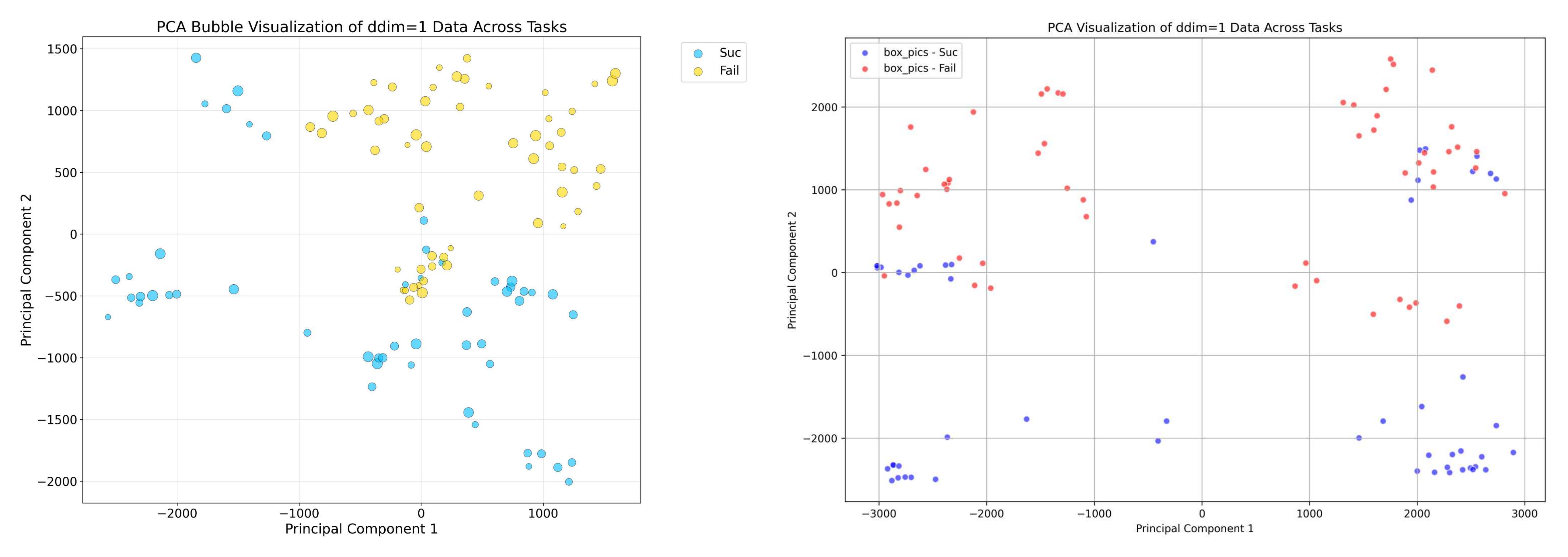}
    \caption{We applied the PCA analysis to the latent single-step feature(left), and LoDiff generated video features(right). }
    \vspace{-1em}
    \label{fig:app_pca}
\end{figure*}

\paragraph{Visual Feature Analysis via PCA}
To better understand how early-stage video predictions correlate with task outcomes, we perform a PCA-based analysis on DDIM=1 predictions. Specifically, we extract the lower half of each generated image (corresponding to the DDIM step 1 visualization), segment it into per-frame patches, and analyze their pixel-level statistics. The extracted features are flattened and reduced using Principal Component Analysis (PCA) to identify structural differences between successful and failed trajectories. This provides insight into whether early prediction quality can serve as a proxy for downstream task success. The detailed pipeline is summarized in Table~\ref{tab:image_analysis_pipeline}.

\begin{table}[t]
\centering
\caption{Pipeline for visual feature analysis based on DDIM=1 predictions.}
\label{tab:image_analysis_pipeline}
\renewcommand{\arraystretch}{1.2}
\resizebox{\linewidth}{!}{
\begin{tabular}{clp{9cm}}
\toprule
\textbf{Step} & \textbf{Operation} & \textbf{Description} \\
\midrule
1 & Image Collection & For each task, collect generated video sequences labeled as \texttt{suc} or \texttt{fail}. Each sequence is stored as a single concatenated image. \\
2 & DDIM=1 Extraction & Extract the bottom half of the image, which corresponds to the model's DDIM=1 prediction. \\
3 & Frame Segmentation & Horizontally split the DDIM=1 region into 4 equal parts, each representing a predicted frame. \\
4 & Feature Flattening & Flatten each frame patch into a 1D vector using grayscale pixel values. \\
5 & Label Assignment & Assign success (1) or failure (0) labels based on task outcome; record task name as metadata. \\
6 & PCA Reduction & Apply PCA to reduce high-dimensional pixel vectors to 2D space for visualization. \\
7 & Bubble Plot Visualization & Plot the PCA-reduced features using color-coded bubbles (blue for success, yellow for failure), with randomized sizes and edge highlights. \\
\bottomrule
\end{tabular}
}
\end{table}

\paragraph{Discussion}
As shown in ~\cref{fig:app_pca}, we find that although humans find it easier to tell the difference on LoDiff videos, the HiDiff single-step feature is easier to analyze through an algorithm.

\section{Appendix B: Detailed Model Parameters}

\subsection{Video Generation Model}

Our video generation framework is built upon the Latent Visual Diffusion Model (LVDM) architecture, based on Dynamicrafter~\cite{xing2023dynamicrafter}, with adaptations to suit the low-resolution setting of $128 \times 128$ and short video clips of 4 frames. The model adopts a denoising diffusion process with 1000 steps and a linear noise schedule ranging from $\beta = 0.00085$ to $\beta = 0.012$. We use a velocity-based parameterization and enable zero-SNR rescaling to stabilize training.

The generative model consists of three main components: a 3D U-Net backbone, a latent-space autoencoder, and a hybrid conditioning mechanism that incorporates both textual and visual inputs. The U-Net is configured with 4 spatial resolution levels and channel multipliers \{1, 2, 4, 4\}, starting from a base width of 320. It applies spatial attention at resolutions \{4, 2, 1\} and temporal attention over 4-frame sequences. Each level contains two residual blocks, and attention heads have 64 channels. Temporal reasoning is handled via temporal convolution and attention mechanisms, and the model supports both image and text cross-attention.
\begin{table*}[h]
\centering
\caption{Video Generation Model Architecture Overview}
\label{tab:model_architecture}
\begin{tabular}{ll}
\toprule
\textbf{Component} & \textbf{Configuration} \\
\midrule
\textbf{Input Resolution} & $128 \times 128$, 4 frames \\
\textbf{Latent Space} & $16 \times 16 \times 4$ (spatial $\times$ temporal), 4 channels \\
\textbf{Diffusion Steps} & 1000 \\
\textbf{Noise Schedule} & Linear ($\beta \in [0.00085, 0.012]$) \\
\textbf{Parameterization} & Velocity-based \\
\textbf{Conditioning} & Hybrid (Text + Image) \\
\textbf{Text Encoder} & Frozen OpenCLIP (penultimate layer) \\
\textbf{Image Encoder} & Frozen OpenCLIP Image Embedder \\
\textbf{Image Projector} & Transformer Resampler (4 layers, 12 heads, 1024 dim) \\
\bottomrule
\end{tabular}
\end{table*}
The first-stage autoencoder encodes RGB video frames into a latent space of size $16 \times 16 \times 4$ (spatial $\times$ temporal), with 4 channels. The encoder uses a ResNet-style convolutional structure with 4 downsampling levels, each with 2 residual blocks. Attention is not used in the autoencoder, and reconstruction is performed using a symmetric decoder.

For conditioning, a hybrid strategy is adopted. Textual inputs are encoded using a frozen OpenCLIP model (penultimate layer), while visual inputs are processed using a frozen OpenCLIP image encoder followed by a trainable Resampler module. The Resampler is a Transformer-based attention module with 4 layers, 12 heads (each of width 64), and 16 learnable queries. It projects 1280-dimensional OpenCLIP features into a 1024-dimensional conditioning vector that aligns with the U-Net’s context embedding space.

Training is performed on a dataset of 4-frame video clips with spatial resolution $128 \times 128$, sampled at a frame stride of 4. We use a batch size of 8 and train with mixed precision (FP16) across 4 GPUs. Gradients are accumulated every 2 steps and clipped to a norm value of 0.5. The model is initialized from a pretrained checkpoint (\texttt{epoch=53-step=12000.ckpt}) and trained with a base learning rate of $2 \times 10^{-5}$. During training, image samples are periodically generated using DDIM sampling with 50 steps and a guidance scale of 7.5 for qualitative evaluation.

A summary of the model architecture and key parameters is provided in Table~\ref{tab:model_architecture} and Table~\ref{tab:unet_autoencoder}.

\begin{table}[h]
\centering
\caption{Detailed Configuration of U-Net and Autoencoder}
\label{tab:unet_autoencoder}
\begin{tabular}{lll}
\toprule
\textbf{Module} & \textbf{Parameter} & \textbf{Value} \\
\midrule
\multirow{6}{*}{\textbf{U-Net}} 
& Base Channels & 320 \\
& Channel Multipliers & \{1, 2, 4, 4\} \\
& Attention Resolutions & \{4, 2, 1\} \\
& Transformer Depth & 1 \\
& Head Channels & 64 \\
& Temporal Length & 4 \\
\midrule
\multirow{5}{*}{\textbf{Autoencoder}} 
& Input Channels & 3 \\
& Embedding Dim & 4 \\
& Resolution & $128 \times 128$ \\
& Channel Multipliers & \{1, 2, 4, 4\} \\
& ResBlocks per Level & 2 \\
\bottomrule
\end{tabular}
\end{table}

\subsection{DiT-based Action Model Architecture}

Our action model is based on the Diffusion Transformer (DiT) architecture~\cite{Peebles2022DiT}, originally designed for image generation tasks, and we extend it for temporally-conditioned action prediction in robotics. The model is adapted to predict future robot actions using a diffusion process, with cross-attention to visual features and classifier-free guidance.

\paragraph{Architecture Overview.}
The action model receives as input: (1) a noisy action sequence $x \in \mathbb{R}^{N \times T \times D}$, (2) a diffusion timestep $t$, and (3) condition embeddings $z$, including both task embeddings and visual features. The architecture consists of four key components:

Action Embedder: A linear projection maps the input action sequence to the transformer hidden space.

Timestep Embedder: A sinusoidal MLP-based embedding module encodes the scalar timestep $t$ into a vector.

Condition Embedder: A $LabelEmbedder$ projects task condition features and applies stochastic token dropping to support classifier-free guidance~\cite{ho2022classifier}.

Transformer Backbone: A stack of $L$ DiT blocks (each with self-attention, optional cross-attention, and MLP sublayers) processes the embedded sequence.

\paragraph{Cross-Attention with Video Features.}
Each DiT block optionally includes a cross-attention mechanism that injects video features (extracted from a pretrained encoder) as conditioning. A residual gate balances the contribution of cross-attention dynamically:

\[
x \leftarrow x + \tanh(\gamma) \cdot \text{CrossAttn}(\text{LN}(x), \text{video\_features})
\]

where $\gamma$ is a learnable gating parameter initialized to 1.

\paragraph{Prediction Head.}
The final prediction is computed via two linear layers:

- One for the action outputs (e.g., joint positions or velocities),

- One for the gripper state (binary open/close), processed separately to improve stability.

These outputs are concatenated and returned as the model prediction for each timestep.

\paragraph{Classifier-Free Guidance.}
To enable conditional generation control, we implement classifier-free guidance by duplicating the batch and replacing condition tokens with a learned null token in the unconditional path. At inference, both conditional and unconditional predictions are combined:

\[
\epsilon_{\text{cfg}} = \epsilon_{\text{uncond}} + s \cdot (\epsilon_{\text{cond}} - \epsilon_{\text{uncond}})
\]

where $s$ is the guidance scale.

\paragraph{Model Variants.}
We consider two model variants:
\begin{itemize}
    \item \textbf{DiT-Small (DiT-S)}: 6 layers, hidden size 384, 4 attention heads.
    \item \textbf{DiT-Base (DiT-B)}: 12 layers, hidden size 768, 12 attention heads.
\end{itemize}

An overview of the configuration is provided in Table~\ref{tab:dit-config}.

\begin{table}[h]
\centering
\caption{DiT Action Model Configuration}
\label{tab:dit-config}
\begin{tabular}{lccc}
\toprule
\textbf{Model} & \textbf{Depth} & \textbf{Hidden Size} & \textbf{Heads} \\
\midrule
DiT-S & 6 & 384 & 4 \\
DiT-B & 12 & 768 & 12 \\
\bottomrule
\end{tabular}
\end{table}

\paragraph{DiffusionMatcher: Temporal Conditioning Module}
\label{sec:diffmatcher}

To bridge the gap between the slow video generation pathway and the fast action diffusion policy, we introduce \textbf{DiffusionMatcher}, a learnable temporal transformer module. It conditions the high-frequency action generation on low-cost video features extracted from intermediate steps of the video diffusion process. 

Specifically, DiffusionMatcher takes a short video clip and a diffusion time step as input, and outputs temporally-aware feature tokens that guide the downstream action model. The design comprises four main components:

\begin{itemize}
    \item \textbf{Frame Embedding:} Each frame is flattened and projected to a hidden space via a two-layer MLP.
    \item \textbf{Temporal Transformer:} A multi-layer Transformer encoder processes frame-wise features to capture temporal dependencies.
    \item \textbf{Time Embedding:} The diffusion step is embedded and added to all frame features to provide time-awareness.
    \item \textbf{Cross-Attention with Learnable Tokens:} A set of learnable projection tokens attend to the temporal features to extract condensed, task-relevant representations.
\end{itemize}

This dual-pathway design allows the fast policy to query predictive video features efficiently, enabling real-time control while retaining the benefits of generative modeling.

\begin{table*}[t]
\centering
\caption{Architecture of \textbf{DiffusionMatcher}. $B$: batch size, $C$: channels, $T$: frames, $H \times W$: spatial resolution, $d_h$: hidden size, $d_p$: projection dim.}
\label{tab:diffmatcher_arch}
\renewcommand{\arraystretch}{1.2}
\resizebox{\linewidth}{!}{
\begin{tabular}{llccp{5.5cm}}
\toprule
\textbf{Module} & \textbf{Layer} & \textbf{Input Size} & \textbf{Output Size} & \textbf{Description} \\
\midrule

\multirow{3}{*}{Frame Embedding} 
  & Linear (1) & $B \times T \times (C\cdot H\cdot W)$ & $B \times T \times d_e$ & Maps flattened frame to embed dim \\
  & ReLU       & $B \times T \times d_e$ & $B \times T \times d_e$ & Non-linearity \\
  & Linear (2) & $B \times T \times d_e$ & $B \times T \times d_h$ & Projects to hidden size \\

\midrule

Position Encoding 
  & Learnable Tensor & $1 \times T \times d_h$ & $1 \times T \times d_h$ & Adds frame position info \\

\midrule

\multirow{2}{*}{Time Embedding} 
  & Embedding + MLP & $B$ & $B \times d_h$ & Encodes diffusion timestep \\
  & Broadcast + Add & $B \times 1 \times d_h$ & $B \times T \times d_h$ & Applied to all frames \\

\midrule

Temporal Transformer 
  & $N$ Encoder Layers & $B \times T \times d_h$ & $B \times T \times d_h$ & Captures temporal features \\

\midrule

Projection Tokens 
  & Learnable + Linear & $1 \times T \times d_p$ & $B \times T \times d_p$ & Learnable query tokens \\

\midrule

Cross-Attention 
  & Multihead Attention & Q: $B \times T \times d_p$ \newline K/V: $B \times T \times d_h$ & $B \times T \times d_p$ & Tokens attend to temporal features \\

\bottomrule
\end{tabular}
}
\end{table*}

\paragraph{Loss Function.}
The model is trained using denoising score matching. Given predicted noise $\hat{\epsilon}$ and true noise $\epsilon$ added to the actions, the objective is:

\[
\mathcal{L}_{\text{MSE}} = \mathbb{E} \left[ \left\| \hat{\epsilon} - \epsilon \right\|^2 \right]
\]

This aligns with the standard diffusion model training paradigm, adapted for continuous-valued action spaces.

\section{Appendix C: RLBench Evaluation Setup}
\begin{figure}
    \centering
    \includegraphics[width=0.9\linewidth]{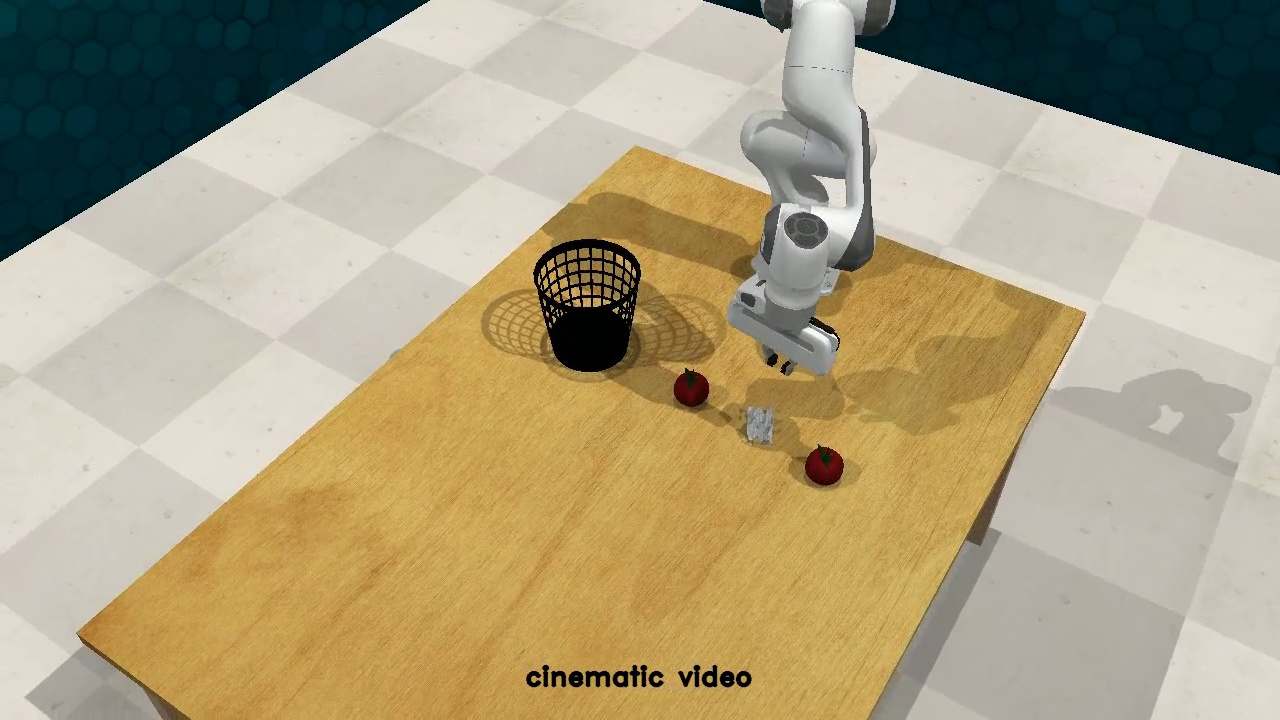}
    \caption{RLBench setup.}
    \label{fig:app_rlbench}
\end{figure}

To evaluate our method in simulation, we adopt the RLBench benchmark~\cite{james2020rlbench}, a large-scale vision-based robot learning environment built on CoppeliaSim. We follow the common protocol for few-shot imitation tasks and evaluate our framework across multiple challenging manipulation scenarios.

\paragraph{Simulator Environment.}
We use the official RLBench Python API with $JointVelocity + DiscreteGripper$ as the action mode and $Front RGB$ camera view at a resolution of $224 \times 224$. The environment is initialized in headless mode for efficient parallel evaluation. Each episode is capped at 15 steps, and success is determined by environment-defined termination signals or task-specific reward triggers.

\paragraph{Evaluated Tasks.}
We select a diverse subset of 7 RLBench tasks to evaluate the generalization and robustness of our model:
\begin{itemize}
  \item \texttt{close\_box}
  \item \texttt{toilet\_seat\_down}
  \item \texttt{sweep\_to\_dustpan}
  \item \texttt{put\_rubbish\_in\_bin}
  \item \texttt{phone\_on\_base}
  \item \texttt{take\_umbrella\_out\_of\_umbrella\_stand}
  \item \texttt{close\_laptop\_lid}
\end{itemize}

These tasks are carefully chosen to cover diverse skill requirements, including object manipulation, precise placement, and long-horizon planning.

\paragraph{Model Evaluation Protocol.}
For each task, we run 25 test episodes using the $predict$ mode, where the model generates actions given language instruction and visual observation. Our policy model is based on a DiT-B transformer backbone trained with our collaborative diffusion-autoregressive mechanism.

During each episode, the model receives observations from the front camera and the current robot state. It then predicts the next action using the $predict_action$ interface, which internally fuses multi-frame visual information. The resulting action is executed in the RLBench environment, and the outcome is recorded.

\paragraph{Execution Details.}
We parallelize task execution across 8 GPUs using $xvfb-run$ and $CUDA\_VISIBLE\_DEVICES$ to enable headless rendering and maximize throughput. Each task is mapped to a different GPU device for simultaneous evaluation. Video recordings of both predicted actions and cinematic rollouts are saved for post-analysis. Average inference time, environment step latency, and episode success rates are logged for each run.

\paragraph{Reproducibility.}
The evaluation script is implemented in Python and based on a wrapper around RLBench and our visual-language-action (VLA) model. All evaluations are deterministic given the same random seed and dataset, and our codebase supports both offline replay and online prediction modes. We provide full configuration files and checkpoints for reproducibility.

\section{Appendix D: Real-World Franka Setups}

\begin{figure*}[h!]
    \centering
    \begin{subfigure}[b]{0.48\linewidth}
        \centering
        \includegraphics[height=5cm]{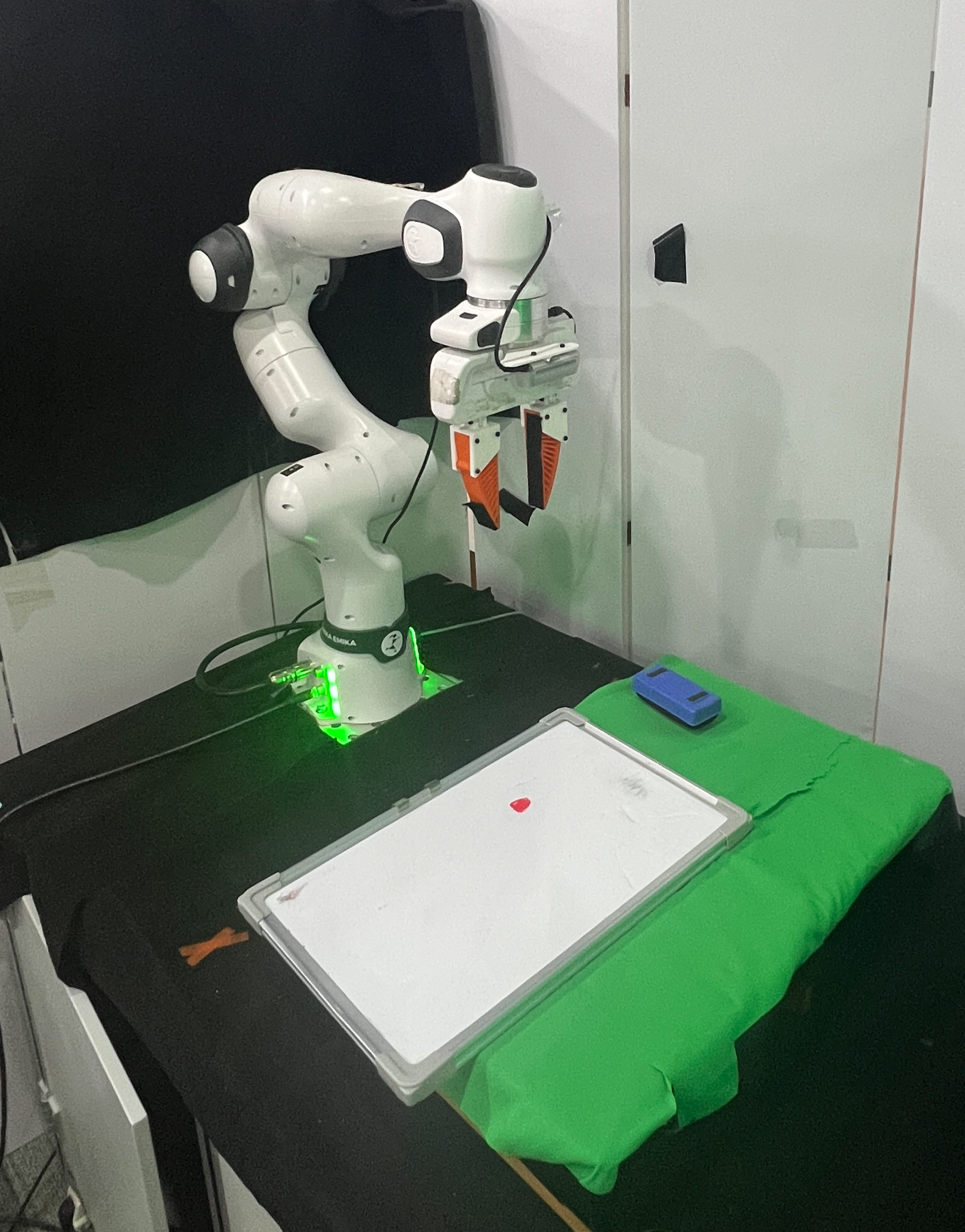}
        \caption{Front view}
        \label{fig:front_view}
    \end{subfigure}
    \hfill
    \begin{subfigure}[b]{0.48\linewidth}
        \centering
        \includegraphics[height=5cm]{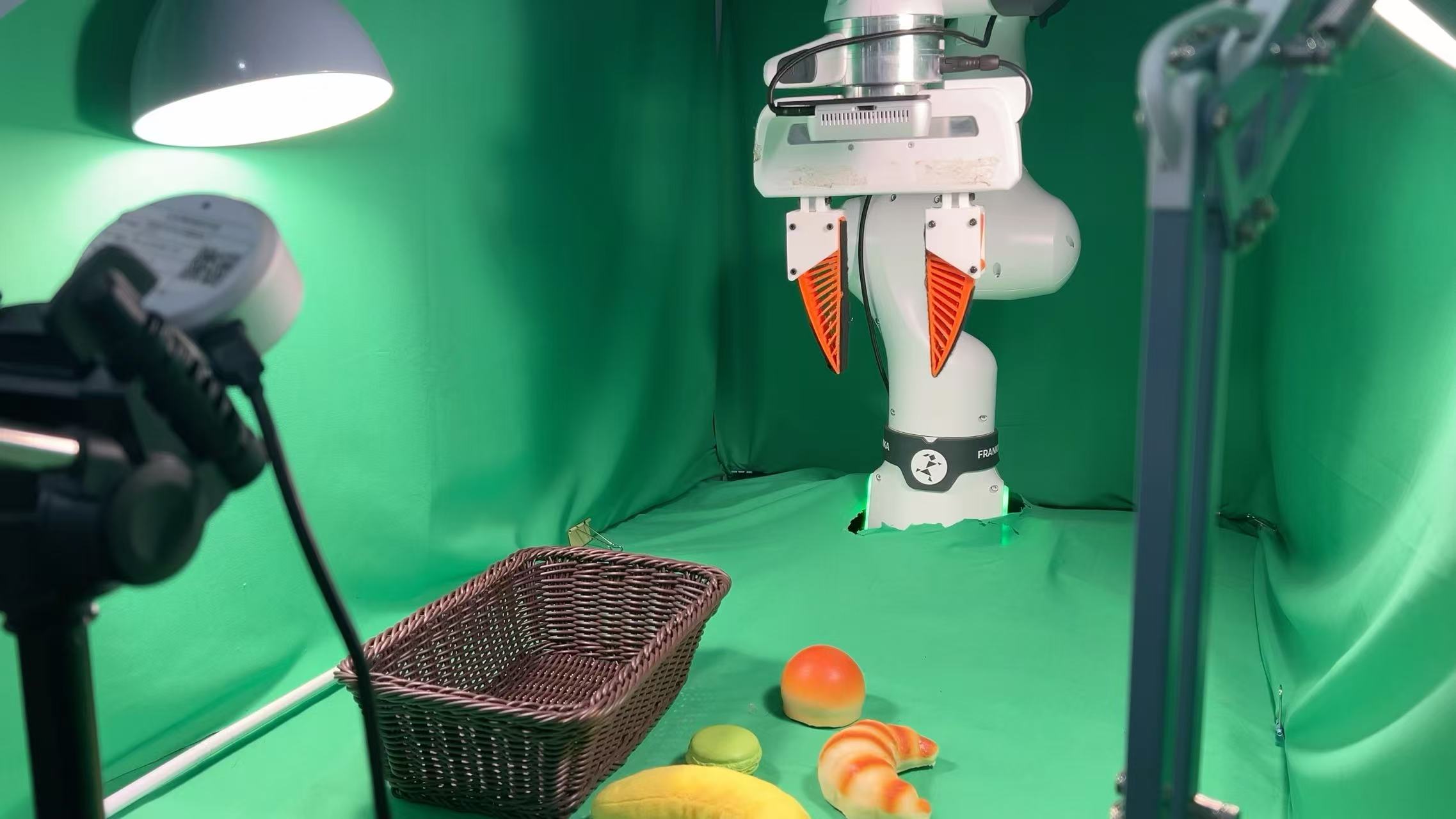}
        \caption{Wrist camera view}
        \label{fig:wrist_view}
    \end{subfigure}
    \caption{Franka Experiment Setups, showing (a) the front view and (b) the wrist camera view.}
    \label{fig:franka_setups}
\end{figure*}

Our real-world robotic experiments are conducted using the Franka Research 3 (FR3) robotic arm, following hardware and software configurations consistent with HybridVLA~\cite{liu2025hybridvla}. Specifically, we operate the FR3 with controller version 5.6.0, \texttt{libfranka} version 0.13.3, and Franka ROS interface version 0.10.0, under Ubuntu 20.04 and ROS Noetic. The robot is set to active execution mode with the FCI switch enabled, allowing low-level torque control and real-time trajectory tracking.

For visual observation, we utilize two Intel RealSense cameras: a front-view \textbf{RealSense D435} and a wrist-mounted \textbf{RealSense D515}, capturing synchronized RGB inputs at a resolution of $224 \times 224$. These inputs are paired with the robot state, which includes full 7-DoF end-effector poses: 3D translation ($\Delta x, \Delta y, \Delta z$), 3D Euler angles (Roll, Pitch, Yaw), and 1D gripper state ($g \in \{0, 1\}$), formulated as:
\[
a = [\Delta x, \Delta y, \Delta z, \text{Roll}, \text{Pitch}, \text{Yaw}, g]
\]

Training demonstrations are collected by teleoperation using a 3Dconnexion SpaceMouse, allowing users to perform tasks from various tabletop positions. Each task includes 100 demonstration trajectories, providing diverse motion patterns and object interactions.

We evaluate our model on the following single-arm real-world manipulation tasks, consistent with prior benchmarks~\cite{liu2025hybridvla}:
\begin{itemize}
  \item \textbf{Pick and Place:} Pick a specifically described object and place it into a matching container.
  \item \textbf{Unplug Charger:} Grasp and remove a charger from a socket with appropriate rotation and lifting.
  \item \textbf{Wipe Blackboard:} Grasp an eraser and remove red markings from a whiteboard surface.
\end{itemize}

This setup ensures compatibility with other state-of-the-art VLA models such as Diffusion Policy~\cite{chi2023diffusion} and CogACT~\cite{li2024cogactfoundationalvisionlanguageactionmodel}, and supports robust benchmarking of language-conditioned visuomotor policies in real-world robotic manipulation.

\section{Appendix E: Code and demo video}
We include some core code and demo videos in the supplement material. We promise the full code will be open-source.

\begin{figure*}[h]
    \centering
    \includegraphics[width=1.\linewidth ]{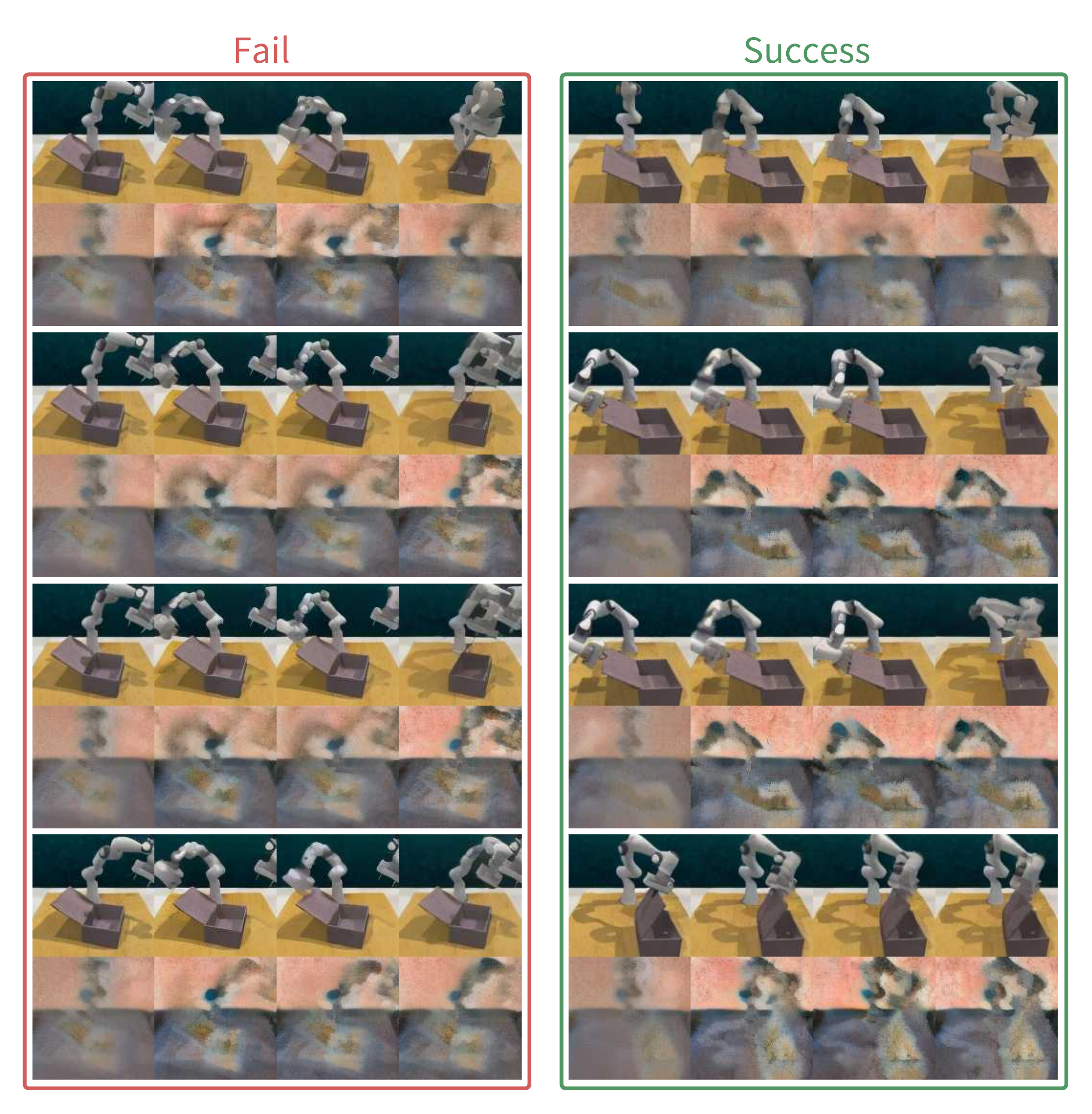}
    \caption{More visualization results of the video generation in feature level or DDIM=100 on RLBench. We select the last four steps of the fail and success cases on the same task. }
    \vspace{-1em}
    \label{fig:app_riskvis}
\end{figure*}


\section{Appendix F: Causal Intervention Experiments}
We designed a series of causal intervention experiments to MinD, showing the affect of the successful rate of different modules.

\begin{table*}[ht]
\centering
\caption{Causal Intervention on MinD Components. We perform controlled interventions on different modules and report the resulting task success rates. A significant drop indicates a causal dependency.}
\label{tab:causal-intervention}
\begin{tabular}{lccc}
\toprule
\textbf{Intervention Type} & \textbf{Module Affected} & \textbf{Success Rate (\%)} & \textbf{$\delta$ vs. Baseline} \\
\midrule
No Intervention (Baseline) & - & 63.0 & - \\
\midrule
\textit{Visual Perturbation} & LoDiff Output ($\hat{v}_t$) & 42.5 & ↓ 20.5 \\
\textit{Latent Masking} & DiffMatcher ($z_t$) & 36.8 & ↓ 26.2 \\
\textit{Random Latent Injection} & HiDiff Condition ($z_t$) & 31.2 & ↓ 31.8 \\
\textit{Frozen LoDiff (No Imagination)} & Whole LoDiff & 28.5 & ↓ 34.5 \\
\textit{LoDiff from Failed Trajectory} & Cross-task Injection & 35.0 & ↓ 28.0 \\
\textit{Swapped Instruction} & Language Input & 30.6 & ↓ 32.4 \\
\bottomrule
\end{tabular}
\end{table*}

These results empirically support that the video prediction module (LoDiff) and the feature adapter (DiffMatcher) are causally influential to the final control outcome. Particularly, visual perturbation and latent injection cause over 30\% performance drop, indicating a strong dependency pathway from imagined visual futures to real-world action decisions.

\end{document}